\documentclass[runningheads]{llncs}

\usepackage{eccv}

\usepackage{eccvabbrv}

\usepackage{graphicx}
\usepackage{booktabs}

\usepackage[accsupp]{axessibility}  
\usepackage{multirow}
\usepackage{stfloats}
\usepackage{colortbl}
\usepackage{makecell}
\usepackage[ruled,linesnumbered]{algorithm2e}

\usepackage{hyperref}

\usepackage{orcidlink}

\begin{document}

\title{
Boosting Transferability in Vision-Language Attacks via Diversification along the Intersection Region of Adversarial Trajectory
}

\titlerunning{Boosting Transferability in Vision-Language Attacks}


\author{Sensen Gao $^{1,3}$ \thanks{co-first authors; $^{\dagger}$ co-corresponding authors} \and
Xiaojun Jia $^{2 * \dagger}$   \and
Xuhong Ren $^2$ \and \\
Ivor Tsang $^{2,3}$ \and
Qing Guo $^{3 \dagger}$ 
}

\authorrunning{S.~Gao et al.}

\institute{Nankai University, Tianjin, China
\and
Nanyang Technological University, Singapore 
\and
 CFAR and IHPC,  Agency for Science, Technology and Research (A*STAR), Singapore\\
}

\maketitle

\begin{abstract}
Vision-language pre-training (VLP) models exhibit remarkable capabilities in comprehending both images and text, yet they remain susceptible to multimodal adversarial examples (AEs).
%
Strengthening attacks and uncovering vulnerabilities, especially common issues in VLP models (\textit{e.g.}, high transferable AEs), can advance reliable and practical VLP models. 
A recent work (\textit{i.e.}, Set-level guidance attack) indicates that augmenting image-text pairs to increase AE diversity along the optimization path enhances the transferability of adversarial examples significantly. 
However, this approach predominantly emphasizes diversity around the online adversarial examples (\textit{i.e.}, AEs in the optimization period), leading to the risk of overfitting the victim model and affecting the transferability.
In this study, we posit that the diversity of adversarial examples towards the clean input and online AEs are both pivotal for enhancing transferability across VLP models. 
Consequently, we propose using diversification along the intersection region of adversarial trajectory to expand the diversity of AEs.
To fully leverage the interaction between modalities, we introduce text-guided adversarial example selection during optimization. 
Furthermore, to further mitigate the potential overfitting, we direct the adversarial text deviating from the last intersection region along the optimization path, rather than adversarial images as in existing methods.
Extensive experiments affirm the effectiveness of our method in improving transferability across various VLP models and downstream vision-and-language tasks. Code is available at \url{https://github.com/SensenGao/VLPTransferAttack}.
%
\end{abstract}
\section{Introduction}
\label{sec:Intro}

\begin{figure}[t]
    \centering
    \includegraphics[width=0.9\linewidth]{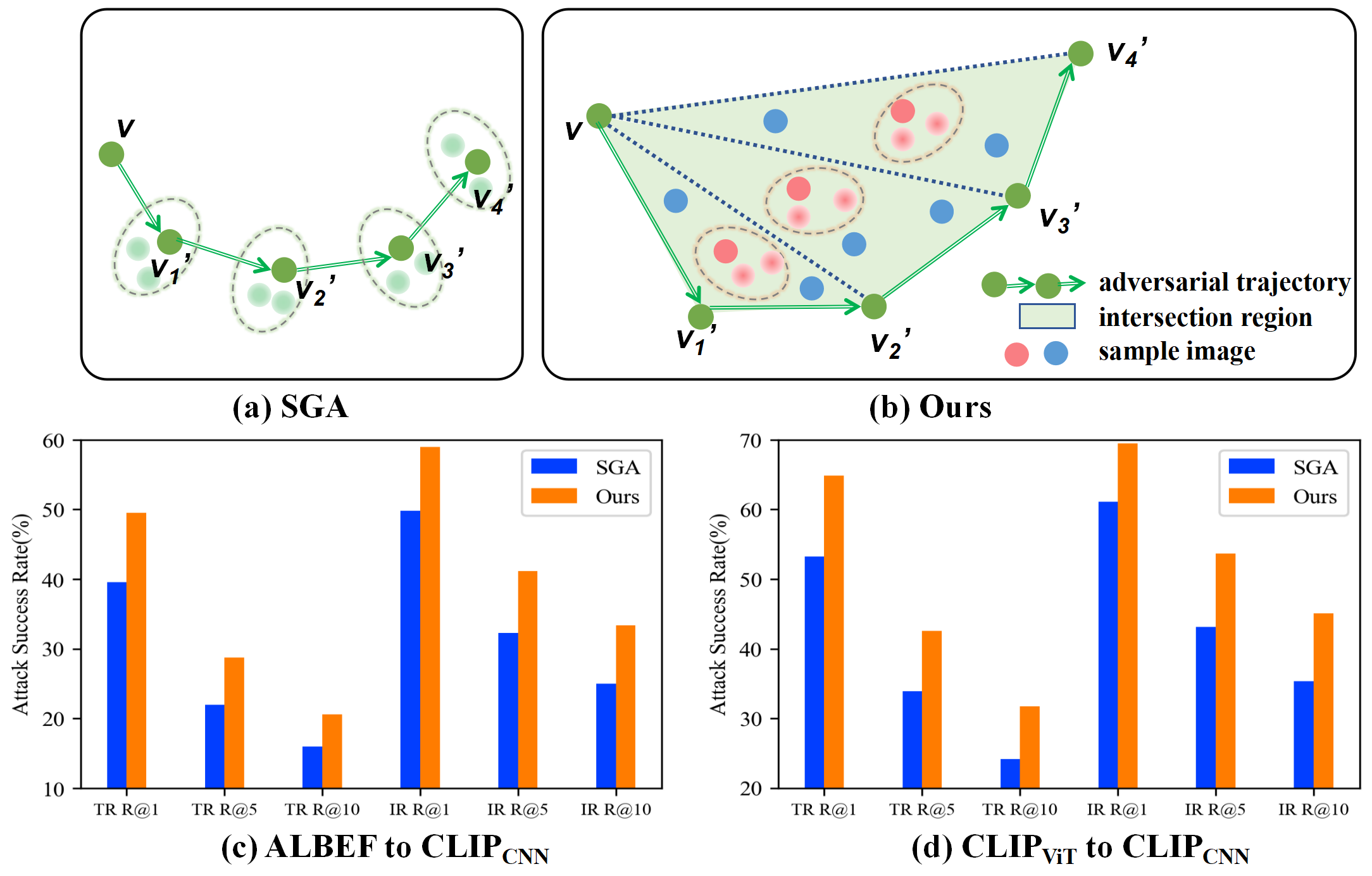}
    \caption{Our method vs. set-level guided attack (SGA) \cite{lu2023setlevel}. (a) shows the main idea of SGA, \textit{i.e.}, conducting augmentation around the online adversarial examples. (b) shows the main idea of our method, that is, we perform augmentation in the intersection region of adversarial trajectory. The red and blue dots both depict images sampled from the intersection region, with red dots indicating the best samples selected using the text-guided adversarial example selection strategy. The surrounding light red dots represent applying the same resizing data augmentation to the best samples as utilized in SGA. (c) and (d) compare the transferability of our method and SGA by using the adversarial examples of ALBEF \cite{li2021align} and CLIP$_\text{ViT}$ to attack CLIP$_\text{CNN}$, respectively.}
    \label{fig:method}
\end{figure}

Vision-language pre-training (VLP) models utilize multimodal learning, leveraging large-scale image-text pairs to bridge the gap between visual and language understanding. These models showcase revolutionary performance across various downstream Vision-and-Language tasks, including image-text retrieval, image captioning, visual grounding, and visual entailment, as demonstrated in \cite{khan2021exploiting, Hu_2022_CVPR,lei2021understanding,shi2021dense}. Despite their success, recent research underscores the significant vulnerability of VLP models, particularly when confronted with multimodal adversarial examples \cite{zhang2022towards,lu2023setlevel,han2023ot,he2023sa,cheng2024typography,luo2024image,gao2024adversarial}.
Uncovering the vulnerabilities, especially common issues, can drive further research aimed at building more reliable and practical VLP models.

Current research predominantly concentrates on attacking VLP models via a white-box setting, where the model's structural information can be exploited. However, exploring the transferability of multimodal adversarial examples is pivotal, especially given the limited access to detailed model structures in real-world scenarios. There have been some efforts made to enhance the transferability of attacks on VLP models by introducing input diversity, as seen in the work SGA \cite{lu2023setlevel}. While they have achieved some effectiveness, how to enhance the transferability of multimodal adversarial examples is still an open question.

In this paper, we undertake a comprehensive examination of the factors contributing to the limited transferability of the cutting-edge multimodal attack method, \textit{i.e.}, SGA \cite{lu2023setlevel}. As shown in Figure \ref{fig:method} (a), throughout the iterative generation of subsequent adversarial images, SGA conducts data augmentation around the online adversarial image (\textit{i.e.}, adversarial examples generated during optimization). This strategy enhances the diversity of adversarial examples along the optimization path, leading to a certain improvement in transferability. 
However, such an approach still carries the potential risk of overfitting to the victim model with the high reliance on the examples along the adversarial trajectory (See Figure \ref{fig:method} (b)), which leads to low attack success rates when we migrate the adversarial example to other VLP models (See Figure \ref{fig:method} (c) and (d)).
To mitigate this overfitting risk, one potential solution is to further enhance the diversity of augmented adversarial examples in a judicious manner.

Building upon the aforementioned analysis, SGA overfits local adversarial examples, while the clean image is the only accessible example that is far from local adversarial examples.
Therefore, we embark on a pioneering endeavor to enhance the transferability of multimodal adversarial attacks by considering the diversity of adversarial examples (AEs) around clean inputs and online AEs throughout the optimization process. To achieve this, we consider the intersection region of adversarial trajectory, which encompasses the original image, the adversarial image from the previous step, and the current adversarial image during the iterative attack process (depicted in Figure \ref{fig:method} (b)). This innovative approach aims to circumvent overfitting by strategic sampling within this region, thereby avoiding an undue focus on adversarial example diversity solely around adversarial images. After obtaining multiple samples, we calculate gradients for each to determine perturbation directions away from the text. Subsequently, we individually incorporate these perturbations into the current adversarial image and select the one that deviates the most from the text.


Additionally, in the text modality, SGA only considers deviating the text from the last adversarial image in the optimization period, but the adversarial image is solely generated by the surrogate model, still posing the risk of overfitting the surrogate model. For this reason, we propose to have the text deviate simultaneously from the last intersection region along the optimization path.

Our proposed method is evaluated on two widely recognized multimodal datasets, Flickr30K \cite{plummer2015flickr30k} and MSCOCO \cite{lin2014microsoft}. We conduct experiments on three vision-and-language downstream tasks (\textit{i.e.}, image-text retrieval (ITR), visual grounding(VG), and image captioning (IC)), and all results indicate the high effectiveness of our method in generating more transferable multimodal adversarial examples. Moreover, when adversarial examples generated from image-text retrieval are transferred to other vision-and-language downstream tasks (\textit{i.e.}, VG and IC), there is a substantial improvement in attack performance.

Our main contributions can be summarized as follows:
\begin{itemize}
    \item We propose using the intersection region of adversarial trajectory to expand the diversity of adversarial examples during optimization, based on which we develop a high-transferability attack against VLP models.
    \item We extend the generation of adversarial text to deviate from the last intersection region along the optimization path, aiming to reduce overfitting the surrogate model, thereby achieving enhanced transferability.
    \item Extensive experiments robustly demonstrate the efficacy of our proposed method in elevating the transferability of multimodal adversarial examples across diverse models and three downstream tasks.
\end{itemize}
\section{Related Work}
\label{sec:Related}

\subsection{Vision-Language Pre-training Models}
VLP models leverage multimodal learning from extensive image-text pairs to improve the performance of various Vision-and-Language (V+L) tasks  \cite{li2022blip}. Early VLP models predominantly depend on pre-trained object detectors for acquiring multimodal representations \cite{chen2020uniter,li2020oscar,zhang2021vinvl,wang2022vlmixer,tan2019lxmert}. Recently, with the advent of end-to-end image encoders like the Vision Transformer (ViT) \cite{dosovitskiy2010image,touvron2021training,yuan2021tokens} offering faster inference speeds, some work propose to use them as substitution for computationally expensive object detectors \cite{dou2022empirical,li2022blip,li2021align,wang2023accelerating,yang2022vision}. 

There are two popular approaches for VLP models in learning vision-language representations: the fused architecture and the aligned architecture. Fused VLP models (\textit{e.g.}, ALBEF \cite{li2021align}, TCL \cite{yang2022vision}), initially employ two separate unimodal encoders to learn features for text and images. Subsequently, a multimodal encoder is utilized to fuse the embeddings of text and images. In contrast, aligned VLP models, exemplified by CLIP \cite{radford2021learning}, focus on aligning the feature spaces of distinct unimodal encoders and benefit downstream tasks significantly \cite{abdelfattah2023cdul}. This paper concentrates on evaluating our proposed method using multiple popular fused and aligned VLP models. 

\subsection{Downstream Vision-and-Language Tasks}
\textbf{Image-Text Retrieval (ITR)} involves retrieving pertinent information, textual or visual, in response to queries from another modality \cite{chen2020imram,cheng2022vista,wang2019camp,zhang2020context}. This undertaking usually encompasses two sub-tasks: image-to-text retrieval (retrieving text based on an image query) and text-to-image retrieval (retrieving images given a text query).

In aligned VLP models, both the Text Retrieval (TR) and Image Retrieval (IR) tasks leverage ranking results determined by the similarity between text and image embeddings. However, in fused VLP models, where internal embedding spaces lack alignment across unimodal encoders, the similarity scores between image and text modalities are computed for all image-text pairs to retrieve the Top-N candidates. Subsequently, these Top-N candidates serve as input for the multimodal encoder, which computes the image-text matching score to establish the final ranking.

\textbf{Visual Grounding (VG)} refers to the task of localizing the region within a visual scene with corresponding entities or concepts in natural language. Among VLP models, ALBEF expands Grad-CAM \cite{selvaraju2017grad} and utilizes the acquired attention map to rank the detected proposals \cite{yu2018mattnet}.

\textbf{Image Captioning (IC)} is to generate a textual description that logically describes or implies the content of a given visual input, typically involving the creation of captions for images. The evaluation of Image Captioning models often employs metrics such as BLEU \cite{naseer2021intriguing}, METEOR \cite{banerjee2005meteor}, ROUGE \cite{lin2004rouge}, CIDEr \cite{vedantam2015cider} and SPICE \cite{anderson2016spice}, which serve to assess the quality and relevance of the generated captions in comparison to reference captions. 

\subsection{Transferability of Adversarial Examples}
Adversarial attacks~\cite{jia2020adv,guo2020abba,guo2021learning,he2023generating,gu2023survey} are typically categorized as white-box and black-box attacks. In a white-box setting~\cite{jia2023improving,huang2023ala}, the attacker has full access to the model, whereas black-box attacks~\cite{bai2020improving,park2024hard}, more realistic in practical applications, occur when information about the model is limited. In the realm of image attacks \cite{huang2024personalization,huang2022ala,huang2021advfilter}, prevalent methods for crafting transferable adversarial examples often leverage data augmentation techniques (\textit{e.g.}, DIM \cite{xie2019improving}, TIM \cite{dong2019evading}, SIM\cite{lin2019nesterov}, ADMIX \cite{wang2021admix}, PAM \cite{zhang2023improving}). Zhang \etal \cite{zhang2022towards} introduced a white-box attack targeting popular VLP models for downstream tasks in the multimodal domain. Building upon this work, Lu \etal \cite{lu2023setlevel} proposed SGA, considering the diversity of adversarial examples by expanding single image-text pairs to sets of images and texts to conduct black-box attacks on VLP models. 

However, SGA primarily emphasizes diversity in the vicinity of adversarial examples during the optimization process, potentially increasing the risk of overfitting the victim model and impacting transferability. Therefore, our primary focus in this study is to further enhance the diversity of adversarial examples in a thoughtful manner and avoid an undue focus on adversarial example diversity solely around adversarial images.

\section{Methodology}
\label{sec:method}

\subsection{Background and Motivation}
\label{subsec:background-motivation}

\begin{table*}[t]
\caption{\textbf{Attack Success Rate (\%) of SGA with and without image augmentation.} The SGA w.o. Aug doesn't consider image augmentation. We use ALBEF to generate multimodal adversarial examples on the ITR task to evaluate transferability.}
\begin{center}
\small
\renewcommand\arraystretch{1}
\setlength{\tabcolsep}{8pt}
    \resizebox{0.9\linewidth}{!}{
		\begin{tabular}{ @{\extracolsep{\fill}} c|c|cc|cc|cc|cc} 
        \toprule[0.3mm]
			& &  \multicolumn{2}{c}{\textbf{ALBEF}} & \multicolumn{2}{c}{\textbf{TCL}} & \multicolumn{2}{c}{\textbf{CLIP$_{\rm ViT}$}} & \multicolumn{2}{c}{\textbf{CLIP$_{\rm CNN}$}}  \\
			\cmidrule{3-10} 
			\multirow{-2}{*}{\textbf{Source}} &\multirow{-2}{*}{\textbf{Attack}} & {TR R@1} & {IR R@1} & {TR R@1} & {IR R@1} & {TR R@1} & {IR R@1} & {TR R@1} & {IR R@1}  \\
			\midrule
			\multirow{2}{*}{\rotatebox[origin=c]{0}{\textbf{ALBEF}}} 
            & SGA w.o. Aug & \textbf{99.9} & 99.95 & 70.07 & 71.67 & 30.55 & 39.88 & 32.31 & 42.54\\
            & SGA w. Aug & \textbf{99.9} & \textbf{99.98} & \textbf{87.88} & \textbf{88.05} & \textbf{36.69} & \textbf{46.78} & \textbf{39.59} & \textbf{49.78}\\
   \bottomrule[0.3mm]
\end{tabular}}
\end{center}
\label{tab:SGAaug}
\end{table*}

Adversarial attacks on VLP models involve inducing a mismatch between adversarial images and corresponding adversarial text while adhering to specified constraints on image and text perturbations. Here, $(v, t)$ denotes an original image-text pair from a multimodal dataset, with $v'$ representing an adversarial image and $t'$ denoting adversarial text. The allowable perturbations are restricted within the ranges $B[v, \xi_{v}]$ for images and $B[t, \xi_{t}]$ for text. The image and text encoders of the multimodal model are denoted as $F_{I}$ and $F_{T}$, respectively. To generate valid multimodal adversarial examples, the objective is to maximize the loss function $J$ specific to VLP models:
\begin{equation}
\label{eq:Problem-Formulation}
\left\{
\begin{array}{ll}
\max J\left(F_{I}(v'),F_{T}(t')\right) \\
\textit{s.t.} v' \in B[v,\xi_{v}], t' \in B[t,\xi_{t}].
\end{array}\right.
\end{equation}

The state-of-the-art approach for exploring the transferability of multimodal adversarial examples (\textit{i.e.}, SGA \cite{lu2023setlevel}) involves augmenting image-text pairs to enhance the diversity of adversarial examples along the optimization path. Specifically, during the iterative generation of adversarial images, let $v'_i$ represent the adversarial image generated at the $i$-th step. In the subsequent step ($i+1$), SGA initiates the process by applying a resizing operation for data augmentation to $v'_i$, resulting in $V'_{i} = \{v'_{i1}, v'_{i2}, ..., v'_{iM}\}$ (See Figure \ref{fig:method} (a)). The iterative formula can be expressed as follows:
\begin{equation}
\label{eq:SGA}
v'_{i+1} = v'_{i} + \alpha \cdot sign (\frac{\nabla_{v} \sum_{j=1}^{M} J(F_{I}(v'_{ij}),F_T(t))}{\Vert\nabla_{v} \sum_{j=1}^{M} J(F_{I}(v'_{ij}),F_T(t))\Vert} ).
\end{equation}

To further examine the impact of image augmentation along the optimization path in the SGA method, we utilize ALBEF as a surrogate model to generate multimodal adversarial examples. These examples are then employed to target VLP models such as TCL and CLIP, assessing the transferability of the attacks. Detailed results are presented in Table \ref{tab:SGAaug}. Our observations reveal that SGA enhances the transferability of adversarial attacks, showing an increase ranging from 6.14\% to 17.81\%.
However, it is noteworthy that the success rate of attacks on the target models remains notably lower than that on the source model. This discrepancy is primarily attributed to the fact that SGA predominantly emphasizes diversity around AE $v'_i$ during the optimization period, without adequately considering the diversity of adversarial examples toward the clean image, leading to the risk of overfitting the victim model and affecting the transferability.

For this purpose, we propose to consider diversification along the intersection region of adversarial trajectory, which encompasses the original image $v$, the adversarial image from the previous step $v'_{i-1}$, and the current adversarial image $v'_i$ during the iterative attack process. This region is established to sample images within it to broaden the diversity of adversarial examples (See Figure \ref{fig:method} (b)). Moreover, to fully leverage the interplay between modalities, we aim for perturbations guided by textual information that induce $v'_i$ to deviate significantly from the associated text $t$. Additionally, in the text modality, our objective is to identify adversarial perturbations that simultaneously deviate from the intersection region rather than only adversarial images, thereby reducing overfitting the surrogate model and enhancing the effectiveness of black-box attacks.

\subsection{Diversification along the Intersection Region}
\label{subsec:Diversification along the Intersection Region}

As outlined in Section \ref{subsec:background-motivation}, we enhance the diversity of adversarial examples by introducing diversification along the intersection region.
Specifically, at the $i$th iteration during optimization, we have the $v'_{i}$, $v'_{i-1}$, and the clean $v$, and these variables form a triangle region denoted as $\triangle v v'_{i-1} v'_{i}$, \textit{i.e.}, the intersection region of adversarial trajectory in Figure \ref{fig:method}.
Then, we initially sample multiple instances within the region $\triangle v v'_{i-1} v'_{i}$, representing the set of samples as $e=\{e_1,e_2,...,e_N\}$. 
Each sample can be expressed as $e_k=\beta \cdot v + \gamma \cdot v'_{i-1} + \eta \cdot v'_{i}$, where $\beta+\gamma +\eta = 1.0$. 
Consequently, we can compute the gradient perturbation for each sample. For the k-th sample, denoted as $e_k$, its gradient perturbation $p_k$ is calculated as follows. In this way, we can get a perturbation set $P=\{p_1,p_2,...,p_N\}$ by
\begin{equation}
\label{eq:sample_perturbation}
p_k = \alpha \cdot sign(\frac{\nabla_{e} J(F_{I}(e_k),F_{T}(t))}{\Vert \nabla_{e} J(F_{I}(e_k),F_{T}(t)) \Vert}).
\end{equation}

\subsection{Text-guided Augmentation Selection}
\label{subsec:text-guided aug}
In Section \ref{subsec:Diversification along the Intersection Region}, a diverse perturbation set $P$ is derived from the intersection region. To harness the full potential of modality interactions, we introduce text-guided augmentation selection to obtain the optimal sample. Specifically, we individually incorporate each element from the perturbation set $P$ into the adversarial image $v'_{i}$. The selection process aims to identify the sample that maximally distances $v'_{i}$ from $t$. This procedure can be represented as:

\begin{equation}
\label{eq:text-guided selection}
m = \underset{p_m \in P}{\arg \max} J(F_{I}(v'_{i}+ p_m),F_{T}(t)).
\end{equation}

At this juncture, $e_m$ represents the selected sample. We employ SGA as our baseline and incorporate the image augmentation methods considered along its optimization path. The chosen optimal sample $e_m$ is resized and expanded into the set $E_m = \{e_{m1},e_{m2},...,e_{mM}\}$. Subsequently, we utilize the expanded set $E_m$ to generate the final adversarial perturbation, yielding $v'_{i+1}$:

\begin{equation}
\label{eq:Ours}
v'_{i+1} = v'_{i} + \alpha \cdot sign(\frac{\nabla_{e} \sum_{j=1}^{M} J(F_{I}(e_{mj}),F_T(t))}{\Vert \nabla_{e} \sum_{j=1}^{M} J(F_{I}(e_{mj},F_T(t))\Vert}).
\end{equation}

\subsection{Adversarial Text deviating from the Intersection Region}
In the text modality, SGA only considers deviating adversarial text from the ultimate adversarial image generated during the iterative optimization process. 
If there are a total of T iterations, The generation of $t'$ only considers deviating from the adversarial image $v'_{T}$. 
The adversarial image $v'_{T}$ is exclusively created by the surrogate model, thereby still presenting the risk of overfitting to the surrogate model. 
However, during the optimization process of the adversarial image, the clean image is entirely independent of the surrogate model.
For this reason, we propose to have the text deviate simultaneously from the last intersection region along the optimization path.
Specifically, the adversarial text deviates from the triangle region constituted by $v$, $v'_{T-1}$ and $v'_{T}$.
\begin{equation}
\label{eq:text-deviate}
t'=\underset{t' \in B\left[t, \epsilon_{t}\right]}{\arg \max} (\lambda \cdot J(F_I(v),F_T(t'))+ \mu \cdot J(F_I(v_{T}'),F_T(t')))+ \nu \cdot J(F_I(v_{T-1}'),F_T(t'))).
\end{equation}
We also set adjustable scaling factors, among which $\lambda + \mu + \nu = 1.0$.

\subsection{Implementation Details}
In the specific process of our attack, we employ an iterative approach. In each iteration, we sample within the intersection region of adversarial trajectory, guided by textual information, to select a sample that maximally deviates the current adversarial image $v'_i$ from the text $t$. Subsequently, we subject this sample to image augmentation processing, calculate gradients to determine the perturbation direction, and overlay it onto the current adversarial image, resulting in $v'_{i+1}$. Through multiple iterative steps, we obtain an adversarial image $v'_{T}$. For the text modality, in contrast to previous methods that solely focus on deviating from the adversarial image $v_{T}'$, our goal is to derive an adversarial text $t'$ that simultaneously deviates from the last intersection region $\triangle v v'_{T-1} v'_{T}$ along the optimization path.

\section{Experiments}
\label{sec:experiments}
In this section, we present experimental evidence demonstrating the enhanced transferability of multimodal examples generated from our proposed method across VLP models and various Vision-and-Language tasks. First, in Section \ref{sec:Experimental Settings}, we introduce the experimental settings, including the popular image-text pair datasets and VLP models we use, as well as restrictions for adversarial attacks. Subsequently, the process of searching for optimal parameters is shown in Section \ref{sec:optimal}. After that, we evaluate cross-model transferability in the context of the image-text retrieval task, as detailed in Section \ref{sec:cross-model}. Following this, in Section \ref{sec:cross-task}, we extend our investigation to the transfer of multimodal adversarial examples generated within the image-text retrieval task to other tasks, aiming to gauge cross-task transferability. Lastly, Section \ref{sec:Ablation} outlines ablation studies.

\subsection{Setups}
\label{sec:Experimental Settings}

\textbf{VLP Models.} In our transferability evaluation experiments across various VLP models, we explore two typical architectures: fused and aligned VLP models. We select CLIP \cite{radford2021learning} for the aligned VLP model. CLIP offers a choice between two distinct image encoders, namely ViT-B/16 \cite{dosovitskiy2010image} and ResNet-101 \cite{He_2016_CVPR}, denoted as CLIP$_{\rm ViT}$ and CLIP$_{\rm CNN}$, respectively. In the case of fused VLP models, we opt for ALBEF \cite{li2021align} and TCL \cite{yang2022vision}. 
%
ALBEF uses a 12-layer ViT-B/16 image encoder and two 6-layer transformers for text and multimodal encoding. TCL shares this architecture but has different pre-training objectives.

\textbf{Datasets.}  In this study, we leverage two widely recognized multimodal datasets, namely Flickr30K \cite{plummer2015flickr30k} and MSCOCO \cite{lin2014microsoft}, for evaluating the image-text retrieval task. The Flickr30K dataset comprises 31,783 images, each accompanied by five captions for annotation. Similarly, the MSCOCO dataset consists of 123,287 images, and approximately five captions are provided for each image. 

Additionally, we employ the RefCOCO+ \cite{yu2016modeling} dataset to assess the Visual Grounding task. RefCOCO+ is a dataset containing 141,564 referring expressions for 50,000 objects within 19,992 MSCOCO images. This dataset serves the purpose of evaluating grounding models by focusing on the localization of objects described through natural language. For another Vision-and-Language task, Image Captioning, we leverage the MSCOCO dataset as well.

\textbf{Adversarial Attack Settings.} 
In our study, we adopt adversarial attack settings of SGA \cite{lu2023setlevel} to ensure a fair comparison. Specifically, we leverage BERT-Attack \cite{li2020bertattack} to craft adversarial texts. The perturbation bound $\xi_{t}$ is set as 1
%
and length of word list $W=10$. PGD \cite{madry2017towards} is employed to get adversarial images and the perturbation bound, denoted as $\xi_{v}$, is set as 8/255. Additionally, iteration steps $T $ is set as 10 and each step size $\alpha = 2/255$. Furthermore, when randomly sampling from the intersection region of adversarial trajectory, we set the number of samples to 5. When generating adversarial texts, we set the three parameters $\lambda,\mu,\nu$ in Equation \ref{eq:text-deviate} to 0.6, 0.2, and 0.2 respectively. The values chosen for these adjustable parameters can be found in Section \ref{sec:optimal}.

\textbf{Evaluation Metrics.} 
The key metric for adversarial transferability is the Attack Success Rate (ASR), which measures the percentage of successful attacks among all generated adversarial examples. A higher ASR indicates more effective and transferable attacks.

\subsection{Optimal Parameters}
\label{sec:optimal}
\begin{table*}[t]
\caption{\textbf{Optimal Parameters:} Attack Success Rate(\%) on different settings, \textbf{Top} for different values of $\lambda , \mu, \nu$ and \textbf{Bottom} for different numbers of samples $N$.}
\begin{center}
\small
\renewcommand\arraystretch{1}
\setlength{\tabcolsep}{8pt}
    \resizebox{0.9\linewidth}{!}{
		\begin{tabular}{@{\extracolsep{\fill}} c|c|cc|cc|cc|cc} 
        \toprule[0.3mm]
			& &  \multicolumn{2}{c}{\textbf{ALBEF}} & \multicolumn{2}{c}{\textbf{TCL}} & \multicolumn{2}{c}{\textbf{CLIP$_{\rm ViT}$}} & \multicolumn{2}{c}{\textbf{CLIP$_{\rm CNN}$}}  \\
			\cmidrule{3-10} 
			\multirow{-2}{*}{\textbf{Source}} &\multirow{-2}{*}{\textbf{Attack}} & {TR R@1} & {IR R@1} & {TR R@1} & {IR R@1} & {TR R@1} & {IR R@1} & {TR R@1} & {IR R@1}  \\
			\midrule
			\multirow{14}{*}{\rotatebox[origin=c]{0}{\textbf{ALBEF}}} 
            & $\boldsymbol{[\lambda,\mu,\nu] = [0.2,0.0,0.8]}$ & \textbf{99.9} & 99.93 & 89.67 & 90.5 & 42.21 & 52.0 & 46.23 & 55.44 \\
            & $\boldsymbol{[\lambda,\mu,\nu] = [0.2,0.2,0.6]}$ & \textbf{99.9} & 99.93 & 90.41 & 90.43 & 41.96 & 51.87 & 46.49 & 55.3\\
            & $\boldsymbol{[\lambda,\mu,\nu] = [0.2,0.4,0.4]}$ & \textbf{99.9} & 99.93 & 90.31 & 90.43 & 41.96 & 51.87 & 45.34 & 54.82\\
            & $\boldsymbol{[\lambda,\mu,\nu] = [0.2,0.6,0.2]}$ & \textbf{99.9} & \textbf{99.95} & 90.52 & 90.57 & 42.21 & 51.84 & 45.08 & 54.92\\
            & $\boldsymbol{[\lambda,\mu,\nu] = [0.2,0.8,0.0]}$ & \textbf{99.9} & 99.93 & 90.31 & 90.57 & 41.72 & 51.74 & 46.1 & 54.68\\
            & $\boldsymbol{[\lambda,\mu,\nu] = [0.4,0.0,0.6]}$ & \textbf{99.9} & 99.93 & 91.25 & 90.88 & 45.52 & 55.32 & 48.91 & 57.63 \\
            & $\boldsymbol{[\lambda,\mu,\nu] = [0.4,0.2,0.4]}$ & \textbf{99.9} & 99.93 & 91.25 & 90.83 & 45.03 & 55.22 & 48.28 & 57.77\\
            & $\boldsymbol{[\lambda,\mu,\nu] = [0.4,0.4,0.2]}$ & \textbf{99.9} & 99.93 & 91.15 & 90.71 & 44.91 & 55.28 & 48.15 & 57.87\\
            & $\boldsymbol{[\lambda,\mu,\nu] = [0.4,0.6,0.0]}$ & \textbf{99.9} & 99.93 & 91.15 & 90.88 & 44.91 & 54.77 & 49.04 & 57.56\\
            & $\boldsymbol{[\lambda,\mu,\nu] = [0.6,0.0,0.4]}$ & \textbf{99.9} & 99.93 & 91.46 & 90.95 & \textbf{46.38} & 56.38 & 49.04 & \textbf{59.11}\\
            & \cellcolor{gray! 40} $\boldsymbol{[\lambda,\mu,\nu] = [0.6,0.2,0.2]}$ & \cellcolor{gray! 40} \textbf{99.9} & \cellcolor{gray! 40} 99.93 & \cellcolor{gray! 40} \textbf{91.57} & \cellcolor{gray! 40} \textbf{91.17} &\cellcolor{gray! 40} 46.26 & \cellcolor{gray! 40} \textbf{56.8} & \cellcolor{gray! 40} \textbf{49.55} & \cellcolor{gray! 40} 59.01\\
            & $\boldsymbol{[\lambda,\mu,\nu] = [0.6,0.4,0.0]}$ & \textbf{99.9} & 99.93 & 90.94 & 90.98 & 46.01 & 56.72 & \textbf{49.55} & 58.87\\
            & $\boldsymbol{[\lambda,\mu,\nu] = [0.8,0.0,0.2]}$ & \textbf{99.9} & 99.93 & 90.73 & 90.98 & 46.13 & 56.71 & 49.46 & 58.74\\
            & $\boldsymbol{[\lambda,\mu,\nu] = [0.8,0.2,0.0]}$ & \textbf{99.9} & 99.93 & 90.31 & 90.95 & 45.77 & 56.65 & 49.34 & 58.87\\
 			\midrule
            \multirow{5}{*}{\rotatebox[origin=c]{0}{\textbf{ALBEF}}} 
            & $\boldsymbol{N = 3}$ &  \textbf{99.9} & \textbf{99.95} & 90.62 & 90.79 & 45.64 & 56.54 & 48.83 & 58.73\\
            & $\boldsymbol{N = 4}$ &  99.79 & 99.91 & 91.36 & \textbf{91.17} & 45.83 & 56.78 & 50.45 & 59.01\\
            & \cellcolor{gray! 40} $\boldsymbol{N = 5}$ &  \cellcolor{gray! 40} \textbf{99.9} & \cellcolor{gray! 40} 99.93 & \cellcolor{gray! 40} \textbf{91.57} & \cellcolor{gray! 40} \textbf{91.17} &\cellcolor{gray! 40} \textbf{46.26} & \cellcolor{gray! 40} 56.8 & \cellcolor{gray! 40} 49.55 & \cellcolor{gray! 40} 59.01\\
            & $\boldsymbol{N = 6}$ & \textbf{99.9} & 99.93 & 90.94 & 90.38 & 45.79 & \textbf{56.96} & \textbf{50.7} & 58.52 \\
            & $\boldsymbol{N = 7}$ &  \textbf{99.9} & 99.91 & 89.88 & 90.95 & 45.4 & 56.35 & \textbf{50.7} & \textbf{59.07}\\
			\bottomrule[0.3mm]
        \end{tabular}}
\end{center}
\label{tab:optimal_params}
\end{table*}

\begin{figure}[t]
    \centering
    \includegraphics[width=0.9\linewidth]{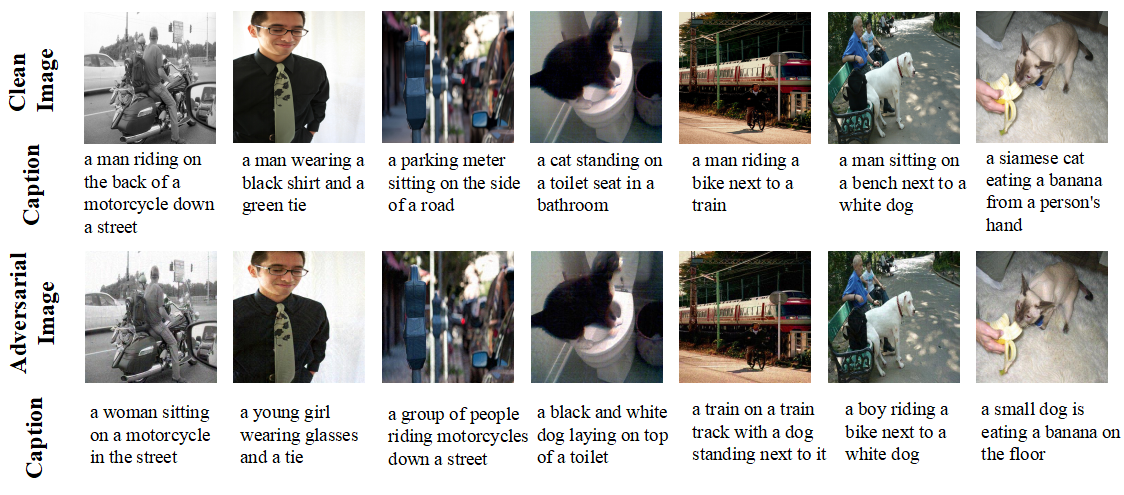}
    \caption{ \textbf{Visualization on Image Captioning.} We use the ALBEF model, pre-trained on Image Text Retrieval(ITR) task, to generate adversarial images on the MSCOCO dataset and use the BLIP \cite{li2022blip} model for Image Captioning on both clean images and adversarial images, respectively.}
    \label{fig:IC_Visualization}
\end{figure}

In our proposed method, the number of samples $N$ taken from the intersection region of adversarial trajectory and scaling factors in Equation \ref{eq:text-deviate} are adjustable. We conduct specific experiments to explore optimal parameter settings and examine their influence on the efficacy of our approach. To be more specific, we utilize ALBEF for generating multimodal adversarial examples on the Flickr30K dataset and assess the transferability on the other three VLP models.

\textbf{Number of Samples taken from Intersection Region of Adversarial trajectory $N$.} 
The bottom of Table \ref{tab:optimal_params} illustrates when the sample size reaches 5, transfer effects are observed for both the Image Retrieval task and the Text Retrieval task. As the sample size continues to increase, the transferability only fluctuates and does not exhibit further improvement. Taking into account both transfer effects and computational costs, a sample size of 5 is identified as the optimal configuration.

\textbf{Scaling factors $\lambda, \mu, \nu$ in Adversarial Text Generation}. In Equation \ref{eq:text-deviate}, $\lambda$ represents the weight of clean images, while $\mu$ and $\nu$ represent the weights of adversarial images. Therefore, we stipulate that $\lambda$ cannot be zero, and $\mu$ and $\nu$ can have at most one zero value. Initially, as the value of $\lambda$ gradually increases, indicating the gradual introduction of clean images, the transferability of multimodal adversarial examples increases accordingly. However, when the value of $\lambda$ becomes too large, it also leads to a disproportionately low proportion of adversarial images, resulting in a decrease in adversarial transferability. According to the experiments, the optimal parameters we select are $[0.6, 0.2, 0.2]$.

\subsection{Cross-Model Transferability}
\label{sec:cross-model}
\begin{table*}[t]
\caption{\textbf{Comparison with state-of-the-art methods on image-text retrieval.} The source column shows VLP models we use to generate multimodal adversarial examples. The gray area represents adversarial attacks under a white-box setting, the rest are black-box attacks. For both Image Retrieval and Text Retrieval, we provide R@1 attack success rate(\%).}
\begin{center}
\small
\renewcommand\arraystretch{1}
\setlength{\tabcolsep}{8pt}
    \resizebox{0.9\linewidth}{!}{
		\begin{tabular}{ @{\extracolsep{\fill}} c|c|cc|cc|cc|cc} 
        \toprule[0.3mm]
			& &  \multicolumn{2}{c}{\textbf{ALBEF}} & \multicolumn{2}{c}{\textbf{TCL}} & \multicolumn{2}{c}{\textbf{CLIP$_{\rm ViT}$}} & \multicolumn{2}{c}{\textbf{CLIP$_{\rm CNN}$}}  \\
			\cmidrule{3-10} 
			\multirow{-2}{*}{\textbf{Source}} &\multirow{-2}{*}{\textbf{Attack}} & {TR R@1} & {IR R@1} & {TR R@1} & {IR R@1} & {TR R@1} & {IR R@1} & {TR R@1} & {IR R@1}  \\
			\midrule
			\multirow{6}{*}{\rotatebox[origin=c]{0}{\textbf{ALBEF}}} 
            & PGD & \cellcolor{gray! 40} 93.74 & \cellcolor{gray! 40} 94.43 & 24.03 & 27.9 & 10.67 & 15.82 & 14.05 & 19.11 \\
            & BERT-Attack & \cellcolor{gray! 40} 11.57 & \cellcolor{gray! 40} 27.46 & 12.64 & 28.07 & 29.33 & 43.17 & 32.69 & 46.11 \\
            & Sep-Attack & \cellcolor{gray! 40} 95.72 & \cellcolor{gray! 40} 96.14 & 39.3 & 51.79 & 34.11 & 45.72 & 35.76 & 47.92\\
            & Co-Attack & \cellcolor{gray! 40} 97.08 & \cellcolor{gray! 40} 98.36 & 39.52 & 51.24 & 29.82 & 38.92 & 31.29 & 41.99  \\
			& SGA  & \cellcolor{gray! 40} \textbf{99.9} & \cellcolor{gray! 40} \textbf{99.98} & 87.88 & 88.05 & 36.69 & 46.78 & 39.59 & 49.78\\
                & \textbf{Ours} & \cellcolor{gray! 40} \textbf{99.9} & \cellcolor{gray! 40} 99.93 & \textbf{91.57} & \textbf{91.17} & \textbf{46.26} & \textbf{56.8} & \textbf{49.55} & \textbf{59.01}\\
			\midrule 
			\multirow{6}{*}{\rotatebox[origin=c]{0}{\textbf{TCL}}} 
            & PGD  & 35.77 & 41.67 & \cellcolor{gray! 40} 99.37 & \cellcolor{gray! 40} 99.33 & 10.18 & 16.3 & 14.81 & 21.1 \\
            & BERT-Attack & 11.89 & 26.82 & \cellcolor{gray! 40} 14.54 & \cellcolor{gray! 40} 29.17 & 29.69 & 44.49 & 33.46 & 46.07 \\
            & Sep-Attack & 52.45 & 61.44 & \cellcolor{gray! 40} 99.58 & \cellcolor{gray! 40} 99.45 & 37.06 & 45.81 & 37.42 & 49.91\\
            & Co-Attack &  49.84 & 60.36 & \cellcolor{gray! 40} 91.68 & \cellcolor{gray! 40} 95.48 & 32.64 & 42.69 & 32.06 & 47.82\\
			& SGA  & 92.49 & 92.77 & \cellcolor{gray! 40} \textbf{100.0} & \cellcolor{gray! 40} \textbf{100.0} & 36.81 & 46.97 & 41.89 & 51.53\\
            & \textbf{Ours} & \textbf{95.2} & \textbf{95.58} & \cellcolor{gray! 40} \textbf{100.0} & \cellcolor{gray! 40} 99.98 & \textbf{47.24} & \textbf{57.28} & \textbf{52.23} & \textbf{62.23}\\
			\midrule 
			\multirow{6}{*}{\rotatebox[origin=c]{0}{\textbf{CLIP$_{\rm ViT}$}}} 
            & PGD & 3.13 & 6.48 & 4.43 & 8.83 & \cellcolor{gray! 40} 69.33 & \cellcolor{gray! 40} 84.79 & 13.03 & 17.43\\
            & BERT-Attack  & 9.59 & 22.64 & 11.80 & 25.07 & \cellcolor{gray! 40}28.34 & \cellcolor{gray! 40}39.08 & 30.40 & 37.43  \\
            & Sep-Attack & 7.61 & 20.58 & 10.12 & 20.74 & \cellcolor{gray! 40} 76.93 & \cellcolor{gray! 40}  87.44 & 29.89 & 38.32\\
            & Co-Attack & 8.55 & 20.18 & 10.01 & 21.29 & \cellcolor{gray! 40} 78.53 & \cellcolor{gray! 40} 87.5 & 29.5 & 38.49\\
			& SGA  & 22.42 & 34.59 & 25.08 & 36.45 & \cellcolor{gray! 40} \textbf{100.0} & \cellcolor{gray! 40} \textbf{100.0} & 53.26 & 61.1\\
                & \textbf{Ours} & \textbf{27.84} & \textbf{42.84} & \textbf{27.82} & \textbf{44.6} & \cellcolor{gray! 40} \textbf{100.0} & \cellcolor{gray! 40} \textbf{100.0}  & \textbf{64.88} & \textbf{69.5}\\
			\midrule 
			\multirow{6}{*}{\rotatebox[origin=c]{0}{\textbf{CLIP$_{\rm CNN}$}}} 
            & PGD & 2.29 & 6.15 & 4.53 & 8.88 & 5.4 & 12.08 & \cellcolor{gray! 40} 89.78 & \cellcolor{gray! 40} 93.04\\
            & BERT-Attack & 8.86 & 23.27 & 12.33 & 25.48 & 27.12 & 37.44 & \cellcolor{gray! 40}30.40 & \cellcolor{gray! 40}40.10\\
            & Sep-Attack & 9.38 & 22.99 & 11.28 & 25.45 & 26.13 & 39.24 & \cellcolor{gray! 40} 93.61 & \cellcolor{gray! 40} 95.3\\
            & Co-Attack & 10.53 & 23.62 & 12.54 & 26.05 & 27.24 & 40.62 & \cellcolor{gray! 40} 95.91 & \cellcolor{gray! 40} 96.5\\
			& SGA  &  15.64 & 28.6 & 18.02 & 33.07 & 39.02 & 51.45 & \cellcolor{gray! 40} \textbf{99.87} & \cellcolor{gray! 40} \textbf{99.9}\\
            & \textbf{Ours} & \textbf{19.5} & \textbf{34.59} & \textbf{21.6} & \textbf{37.88} & \textbf{48.47} & \textbf{59.12} & \cellcolor{gray! 40} \textbf{99.87} & \cellcolor{gray! 40} \textbf{99.9}\\
			\bottomrule[0.3mm]
	\end{tabular}}
\end{center}
\label{tab:flickr}
\end{table*}

As outlined in Section \ref{sec:Experimental Settings}, our experimental design focuses on assessing the transferability of adversarial examples across two widely adopted VLP model architectures: fused and aligned. There are four VLP models selected, namely: ALBEF, TCL, CLIP$_{\rm ViT}$, CLIP$_{\rm CNN}$. The selected downstream V+L task for evaluation is image-text retrieval. Our approach involves employing one of four models to generate multimodal adversarial examples within the specified parameters of our proposed method. Subsequently, we validate the effectiveness of the generated adversarial examples through a comprehensive set of experiments, encompassing both self-attacks and attacks on the other three models. This evaluation encompasses one white-box attack and three distinct black-box attacks.

We adopt the methodology proposed by SGA \cite{lu2023setlevel} as our baseline, Therefore, the effectiveness of our method is compared against it. Moreover, we also present the effectiveness and transferability results of various other attack methods on VLP models. Table \ref{tab:flickr} provides a comprehensive comparison of these methods on the dataset Flickr30K, More experiments on the MSCOCO dataset are provided in the Appendix. In the comparison, PGD \cite{madry2017towards} is an image-only attack, while Bert-Attack \cite{li2020bertattack} focuses solely on text-based attacks. Sep-Attack involves perturbing text and image separately. Furthermore, Co-Attack takes into account cross-modal interactions, generating an adversarial example for one modality under the guidance of the other modality. SGA, our baseline, expands a single image-text pair into a set of images and a set of texts to enhance diversity.

First, we compare the performance of various methods under white-box attacks, wherein we can leverage the model architecture and interact with the model. It is evident that our method, along with SGA, performs the best in the four white-box attack experiments compared to other methods. Whether TR or IR, the attack success rate at the top-1 rank (R@1) consistently exceeds 99.9\%. Given the already high white-box attack success rate achieved by the SGA method, there is limited room for improvement in our approach within this context. Subsequently, we shift our focus to elucidating the enhancements our method brings to transferability, specifically in the realm of black-box attack performance. We delineate this exploration into two segments, contingent on whether the source model and the target model share the same architecture.

\textbf{Cross-Model Transferability in Same Architecture.} 
ALBEF and TCL both utilize a similar model architecture but differ in their pre-training objectives while maintaining a common fundamental model structure. Therefore, when ALBEF and TCL are employed as target models for each other, the success rate of attacks using multimodal adversarial examples is remarkably high. Notably, the black-box attack success rate of IR reaches 95.58\% when TCL is the target model in our proposed method.
Additionally, the transferability of SGA can reach approximately 90\%, but the room for improvement is very limited, our method has improved compared to SGA, ranging from 2.71\% to 3.69\%.
In contrast, when considering CLIP$_{\rm ViT}$ and CLIP$_{\rm CNN}$—both being aligned VLP models—their image encoders vary significantly, with one utilizing the Vision Transformer and the other employing ResNet-101.
%
Given the significant structural differences between traditional CNNs and Transformers, existing methods show low cross-model transferability, highlighting room for improvement. Our method outperforms SGA by about 7.67\% to 11.62\%.

\textbf{Cross-Model Transferability in Different Architectures.} When fused VLP models (\textit{i.e.}, ALBEF and TCL) are used for generating adversarial examples, our method achieves a significant improvement compared to existing methods, outperforming the current state-of-the-art (SOTA) methods by 9.23\% to 10.7\%. When the fused VLP models are targeted for attack, our method still outperforms all existing methods when generating adversarial examples against the aligned VLP models (\textit{i.e.}, CLIP). In this scenario, the transferability of all methods is relatively low. Compared to our method's baseline SGA, we attain an improvement ranging from 2.74\% to 8.25\%.

\subsection{Cross-Task Transferability}
\begin{figure}[t]
    \centering
    \includegraphics[width=0.9\linewidth]{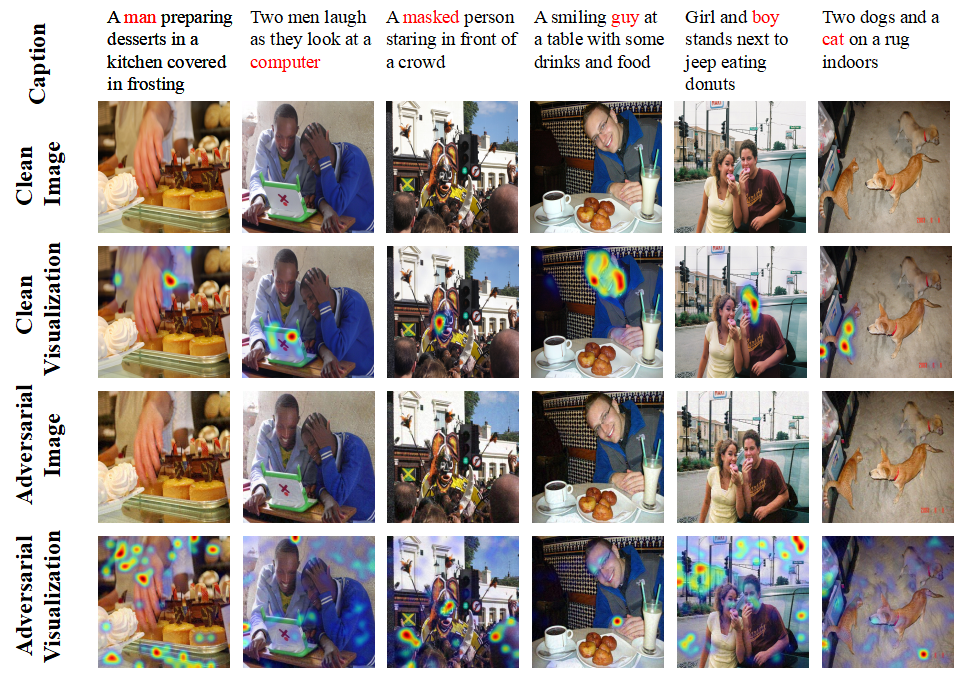}
    \caption{\textbf{Visualization on Visual Grounding.} We use the ALBEF model, pre-trained on the ITR task, to generate adversarial images on the RefCOCO+ dataset and use the same model, pre-trained on Visual Grouding(VG) task, to localize the regions corresponding to red words on both clean images and adversarial images, respectively.}
    \label{fig:VG_Visualization}
\end{figure}

\label{sec:cross-task}
\begin{table*}[t]
\caption{ \textbf{Cross-Task Transferability.} 
We utilize ALBEF to generate multimodal adversarial examples for attacking both Visual Grounding(VG) on the RefCOCO+ dataset and Image Captioning(IC) on the MSCOCO dataset. The baseline represents the performance of each task without any attack, where a lower value indicates better effectiveness of the adversarial attack for both tasks.}
\begin{center}
\small
\renewcommand\arraystretch{1}
\setlength{\tabcolsep}{8pt}
    \resizebox{0.9\linewidth}{!}{
		\begin{tabular}{ @{\extracolsep{\fill}} c|ccc|ccccc} 
        \toprule[0.3mm]
			& \multicolumn{3}{c}{\textbf{ITR $\rightarrow$ VG}} & \multicolumn{5}{c}{\textbf{ITR $\rightarrow$ IC}} \\
			\cmidrule{2-9}
			\multirow{-2}{*}{\textbf{Attack}} & {Val} & {TestA} &    {TestB} & {B@4} & {METEOR} & {ROUGE-L} & {CIDEr} & {SPICE} \\
			\midrule
            Baseline & 58.46 & 65.89 & 46.25 & 39.7 & 31.0 & 60.0 & 133.3 & 23.8\\
            SGA & 50.56 & 57.42 & 40.66 & 28.0 & 24.6 & 51.2 & 91.4 & 17.7 \\
            Ours & \cellcolor{gray! 40} \textbf{49.70}  & \cellcolor{gray! 40} \textbf{56.32} & \cellcolor{gray! 40}\textbf{40.54} & \cellcolor{gray! 40} \textbf{27.2} & \cellcolor{gray! 40} \textbf{24.2} & \cellcolor{gray! 40} \textbf{50.7} & \cellcolor{gray! 40} \textbf{88.3} & \cellcolor{gray! 40} \textbf{17.2}\\
			\bottomrule[0.3mm]
	\end{tabular}}
\end{center}
\label{tab:cross_task}
\end{table*}

We not only assess the transferability of multimodal adversarial examples generated by our proposed method across different VLP models but also conduct experiments to evaluate its effectiveness in transferring across diverse V+L tasks. Specifically, we craft adversarial examples for the Image-Text Retrieval (ITR) task and evaluate them on Visual Grounding (VG) and Image Captioning (IC) tasks. As evident from Table \ref{tab:cross_task} and visual results in Figure \ref{fig:IC_Visualization} and \ref{fig:VG_Visualization}, the adversarial examples generated for ITR demonstrate transferability, successfully impacting both VG and IC tasks. This highlights the efficacy of cross-task transferability in our proposed method. Furthermore, our transferability consistently outperforms that of SGA. 

\subsection{Ablation Study}
\label{sec:Ablation}
Our proposed method builds upon the SGA \cite{lu2023setlevel} as a baseline, introducing two key improvements. Firstly, we utilize diversification along the intersection region of adversarial trajectory to expand the diversity of adversarial examples. Secondly, we generate adversarial text while simultaneously distancing it from the last intersection region along the optimization path. To investigate the impact of each improvement on the effectiveness of our method, we conduct ablation studies on the ITR task and employ transferability from ALBEF to the other three VLP models as the evaluation metric.

\begin{figure}[t]
    \centering
    \includegraphics[width=0.9\linewidth]{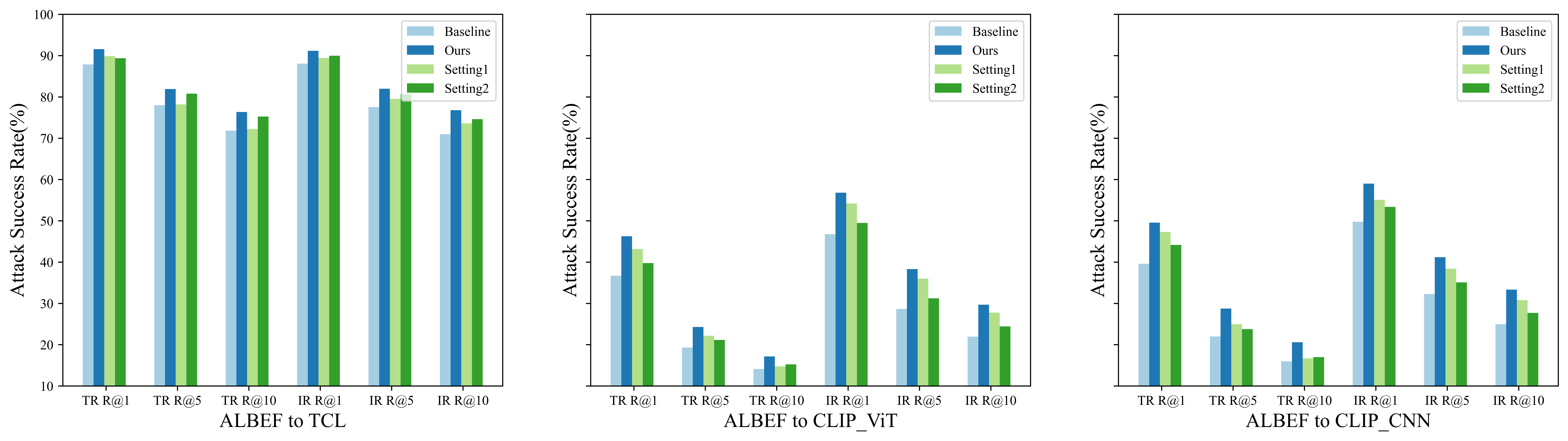}
    \caption{\textbf{Ablation Study: Attack Success Rate(\%) on other three target models.} The baseline is SGA. \textbf{Setting 1} removes diversification along the intersection region of adversarial trajectory.  \textbf{Setting 2} removes the text deviating from the last intersection region along the optimization path.}
    \label{fig:Ablation}
\end{figure}

In the ablation study, we systematically eliminate each enhancement from our approach and compare their transferability with both the baseline SGA and our final method, which incorporates a combination of two improvements. The results are depicted in Figure \ref{fig:Ablation}. It is evident that when attacking models with the same architecture versus different architectures, the impacts of two distinct improvements are also different. When attacking TCL with the same structure as ALBEF, under \textbf{Setting1}, our method exhibits a greater decrease in transferability, indicating that diversification along the intersection region is most crucial in our approach under this scenario. However, under \textbf{Setting2}, the decrease in transferability is more pronounced when attacking the aligned VLP model, indicating that at this point, adversarial texts deviating from the last intersection region play a larger role in the effectiveness of our method.

\section{Conclusion}
In this paper, we conduct a systematic evaluation of existing multimodal attacks regarding transferability. We found that these methods predominantly prioritize diversity around adversarial examples (AEs) during the optimization process, potentially leading to overfitting to the victim model and hindering transferability. To address this issue, we propose diversification along the intersection region of adversarial trajectory to broaden diversity not only around AEs but also towards clean inputs. Moreover, we pioneer an exploration into extending adversarial texts deviating from the intersection region. Through extensive experiments, we demonstrate the effectiveness of our method in enhancing transferability across VLP models and V+L tasks. This work could act as a catalyst for more profound research on the transferability of multimodal AEs, alongside fortifying the adversarial robustness of VLP models.

\clearpage
\section*{Acknowledgments}
This research is supported by the National Research Foundation, Singapore, and DSO National Laboratories under the AI Singapore Programme (AISG Award No: AISG2-GC-2023-008), Career Development Fund (CDF) of Agency for Science, Technology and Research (A*STAR) (No.: C233312028), and National Research Foundation, Singapore and Infocomm Media Development Authority under its Trust Tech Funding Initiative (No. DTC-RGC-04).


%
%
\bibliographystyle{splncs04}
\bibliography{main}

\clearpage
\appendix
\setcounter{figure}{0}
\setcounter{table}{0}
\renewcommand{\thetable}{A\arabic{table}}
\renewcommand{\thefigure}{A\arabic{figure}}

In this appendix, we present our visualizations of multimodal data perturbations and the formulation of attacks. Additionally, we delve into the intersection region of adversarial trajectories and the rationale behind the improvement in transferability. Moreover, we explore adversarial attacks on Large Language Models (LLMs) and compare the time efficiency. Furthermore, we provide more detailed results regarding cross-model transferability.
\begin{itemize}
    \item In Section \ref{sec:visualization}, we present visualizations of the perturbations caused by our multimodal adversarial attack methods on both image and text modalities.
    \item In Section \ref{sec:attack formulation}, we present attack formulation, which outlines the step-by-step process of generating adversarial images and texts.
    \item In Section \ref{sec:optimal sample}, we explore the proportions of clean images, last adversarial images, and adversarial images in the intersection regional optimal samples, as well as the changes in these proportions during the iteration process.
    \item In Section \ref{sec:IIWs}, we discuss the rationale behind the transferability improvement of our method compared to SGA.
    \item In Section \ref{sec:LLMs}, We employ the VLP models as surrogates to generate adversarial examples and explore the adversarial robustness of existing state-of-the-art LLMs.
    \item In Section \ref{sec:time cost}, we compare the time costs between SGA and our method.
    \item In Section \ref{sec:detailed results}, we present more detailed results about cross-model transferability, incorporating the R@5 and R@10 metrics, as well as extending the analysis to the MSCOCO dataset and different perturbation settings.
\end{itemize}

\section{Visualization on Multimodal Dataset}
\label{sec:visualization}
\begin{figure*}[htb]
    \centering
    \includegraphics[width=0.9\linewidth]{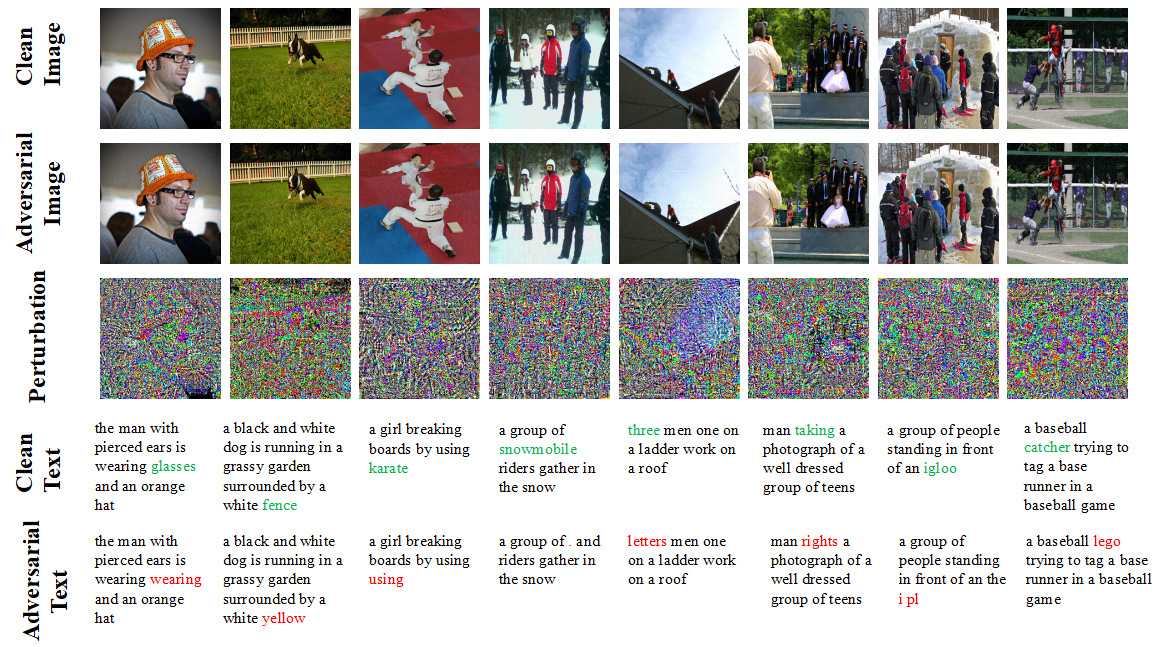}
    \caption{\textbf{Adversarial perturbation visualization on Multimodal dataset.} }
    \label{fig:Flickr_visualization}
\end{figure*}
In Figure \ref{fig:Flickr_visualization}, we present the adversarial perturbations of our method for both image and text modalities. As mentioned in our experimental setup, we perturb only one token for text and use an 8/255 perturbation for images. The first two lines represent the clean image and the adversarial image, respectively. The third line indicates the added adversarial perturbation. The last two lines represent the clean text and the adversarial text, where green highlights the original words, while red highlights the words modified in the adversarial text.
\section{Attack Formulation}
\label{sec:attack formulation}
\begin{algorithm}[t]
\caption{Attack formulation. \label{alg1}}
\LinesNumbered
\KwData{Encoder $F_{I},F_{T}$, perturbation bound $\epsilon_{v},\epsilon_{t}$, Image-caption $(v,t)$, Iteration steps $T$,  Sample number $N$, Augmentation number $M$, Image scale sets $S=\{s_1,s_2,...,s_M\}$}
\KwOut{adversarial image $v'$, adversarial caption $t'$}
Follow SGA method and get adversarial image $v_{0}', v_{1}'$\\
\For{$i = 2, 3, ..., T$}{
$/*$ Build sample set $e=\{e_1,e_2,...,e_N\}$  $*/$ \\
\For{$iter$ $k = 1, 2, ..., N$}{
$e_k=\beta \cdot v + \gamma \cdot v'_{i-1} + \eta \cdot v'_{i},  \beta+\gamma +\eta = 1.0$
}
$/*$ Build perturbation set $P=\{p_1,p_2,...,p_N\}$  $*/$ \\
\For{$iter$ $k = 1, 2, ..., N$}{
$p_k = \alpha \cdot sign(\frac{\nabla_{e} J(F_{I}(e_k),F_{T}(t))}{\Vert \nabla_{e} J(F_{I}(e_k),F_{T}(t)) \Vert}).$
}
$/*$ Get optimal sample $e_m$ $*/$ \\
\quad \space $m = \underset{p_m \in P}{\arg \max} J(F_{I}(v'_{i}+ p_m),F_{T}(t)).$ \\
$/*$ Build augmentation set $E_m = \{e_{m1},e_{m2},...,e_{mM}\}$ $*/$ \\
\For{$iter$ $j = 1, 2, ..., M$}{
$e_{mj} = resize(e_{m}, s_j) + 0.05 \cdot \boldsymbol{N}(0,1)$
}
$/*$ Update adversarial image $v'_{i}$ $*/$ \\
\quad \space $v'_{i+1} = v'_{i} + \alpha \cdot sign(\frac{\nabla_{e} \sum_{j=1}^{M} J(F_{I}(e_{mj}),F_T(t))}{\Vert \nabla_{e} \sum_{j=1}^{M} J(F_{I}(e_{mj},F_T(t))\Vert}).$
}

$/*$ Generate adversarial text $t'$ $*/$ \\
\quad \space $t'=\underset{t' \in B\left[t, \epsilon_{t}\right]}{\arg \max} (\lambda \cdot J(F_I(v),F_T(t'))+ \mu \cdot J(F_I(v_{T}'),F_T(t')))+ \nu \cdot J(F_I(v_{T-1}'),F_T(t'))).$
\end{algorithm}
In Algorithm \ref{alg1}, we provide a detailed demonstration of the process by which our method generates adversarial images and adversarial texts.
\section{Intersection Regional Optimal Sample}
\label{sec:optimal sample}

The current state-of-the-art method, namely SGA (Set-level Guidance Attack) \cite{lu2023setlevel}, focuses on improving the transferability of multimodal adversarial example transferability by exploiting the diversity of adversarial examples along the optimization path. Despite achieving notable success, this approach primarily considers the diversity surrounding adversarial examples within the optimization trajectory. It is crucial to note that adversarial examples generated iteratively by the surrogate model pose a potential risk of overfitting to the victim model, thereby impacting transferability.

To address the potential overfitting risk, we undertake a groundbreaking initiative to enhance the transferability of multimodal adversarial attacks by incorporating diversity not only around online adversarial examples (AEs) throughout the optimization process but also around clean inputs. Specifically, we introduce a novel concept termed the intersection region of adversarial trajectory, which encompasses the original image, the adversarial image from the previous step, and the current adversarial image during the iterative attack process. This pioneering approach aims to mitigate overfitting by strategic sampling within this region, thereby avoiding an exclusive focus on adversarial example diversity solely around adversarial images. Upon acquiring multiple sample points, we compute gradients for each, allowing the determination of a perturbation direction away from the text. Subsequently, we incorporate these perturbations individually into the current adversarial image and select the perturbation that deviates the most from the text.

In the iterative process, the intersection region is defined by the original image $v$, the adversarial image from the previous step $v'_{i-1}$, and the current adversarial image $v'_i$. Each sample within this region can be represented as $e_k = \beta \cdot v + \gamma \cdot v'_{i-1} + \eta \cdot v'_{i}$, where $\beta + \gamma + \eta = 1$. Our objective is to delve into the specific locations within this region where optimal samples occur at each update. This involves a detailed examination of the precise magnitude and variation of the three numerical values, $\beta$, $\gamma$, and $\eta$.

\begin{figure*}[t]
    \centering
    \includegraphics[width=0.9\linewidth]{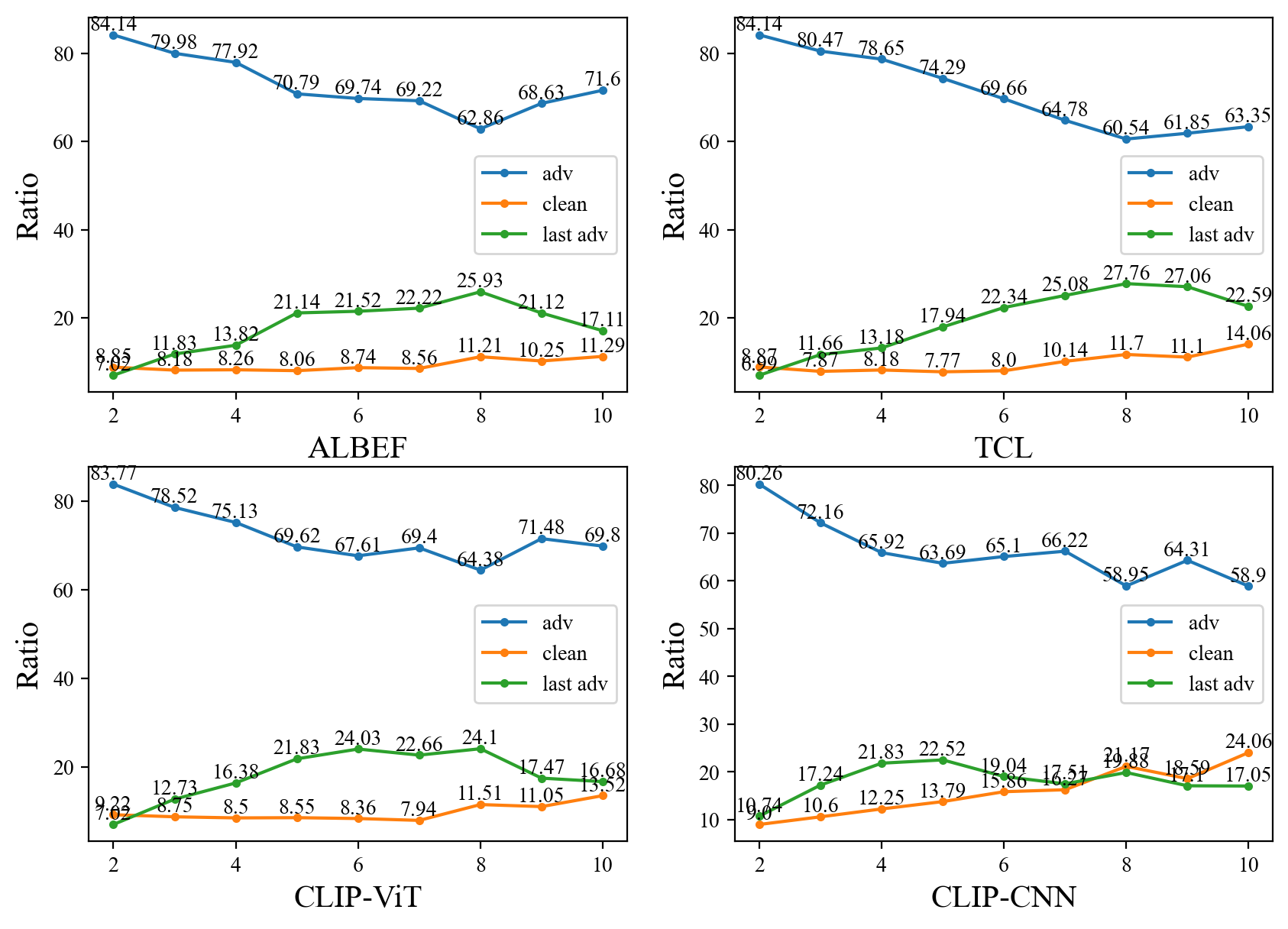}
    \caption{\textbf{Intersection Regional Optimal Sample}. 
    ALBEF, TCL, and CLIP are sequentially used as source models. During the iteration process, the optimal sample from the intersection region of adversarial trajectory is characterized by the proportions of the clean image, the last adversarial image, and the current adversarial image.}
    \label{fig:ratio-final}
\end{figure*}

To achieve the aforementioned objectives, we employ ALBEF \cite{li2021align}, TCL \cite{yang2022vision}, and CLIP \cite{radford2021learning} as surrogate models on the Flickr30K \cite{plummer2015flickr30k} dataset to generate multimodal adversarial examples, respectively. The average value is computed across all examples. Given that we set the PGD \cite{madry2017towards} step size to 10 and utilize random noise initialization, there will be 9 iterations of sampling and selecting the optimal sample. The detailed results are illustrated in Figure \ref{fig:ratio-final}.

First, let's analyze the numerical distribution of the three ratios: $\beta$, $\gamma$, and $\eta$. As depicted in Figure \ref{fig:ratio-final}, it is evident that, for any VLP model and at each step, the proportion of adversarial images consistently remains the highest, typically surpassing 60\%. Furthermore, in the optimal samples, the proportion of adversarial images $v'_{i}$ does not exceed 85\%. In the case of SGA, it can be viewed as a special scenario within our method, wherein the proportion of adversarial images in the selected samples consistently reaches 100\% in each iteration.

Due to an excessive focus on optimizing diversity around adversarial images and insufficient consideration of clean images in the optimization path, the SGA method tends to exhibit overfitting on the surrogate model, thereby limiting its transferability. In contrast, our approach introduces approximately 5-15\% clean images at each iteration while selecting optimal samples, effectively mitigating this risk. Furthermore, within each optimal sample, there exists an additional proportion of 5-25\% derived from the previous adversarial image. This inclusion of historical adversarial perturbations enriches the diversity of adversarial examples, thus effectively preventing the occurrence of overfitting.

\begin{figure*}[t]
    \centering
    \includegraphics[width=0.8\linewidth]{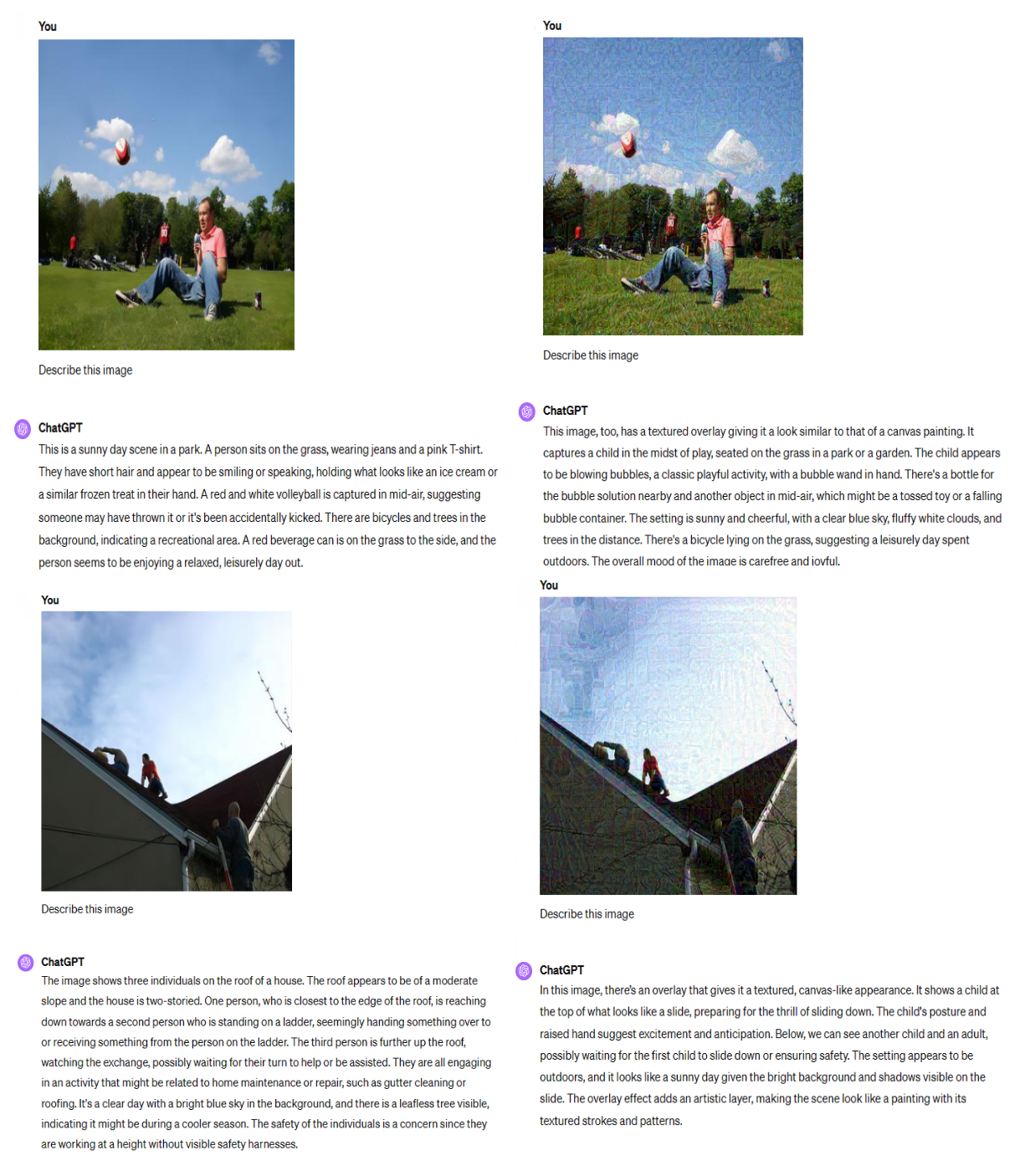}
    \caption{\textbf{Adversarial Transferability on GPT-4.} The images on the left display the responses generated by GPT-4 when presented with \textbf{clean images} and the query "Describe this image", while the images on the right show the responses to the \textbf{adversarial images}.}
    \label{fig:Attack_GPT4}
\end{figure*}

\begin{figure*}[t]
    \centering
    \includegraphics[width=0.8\linewidth]{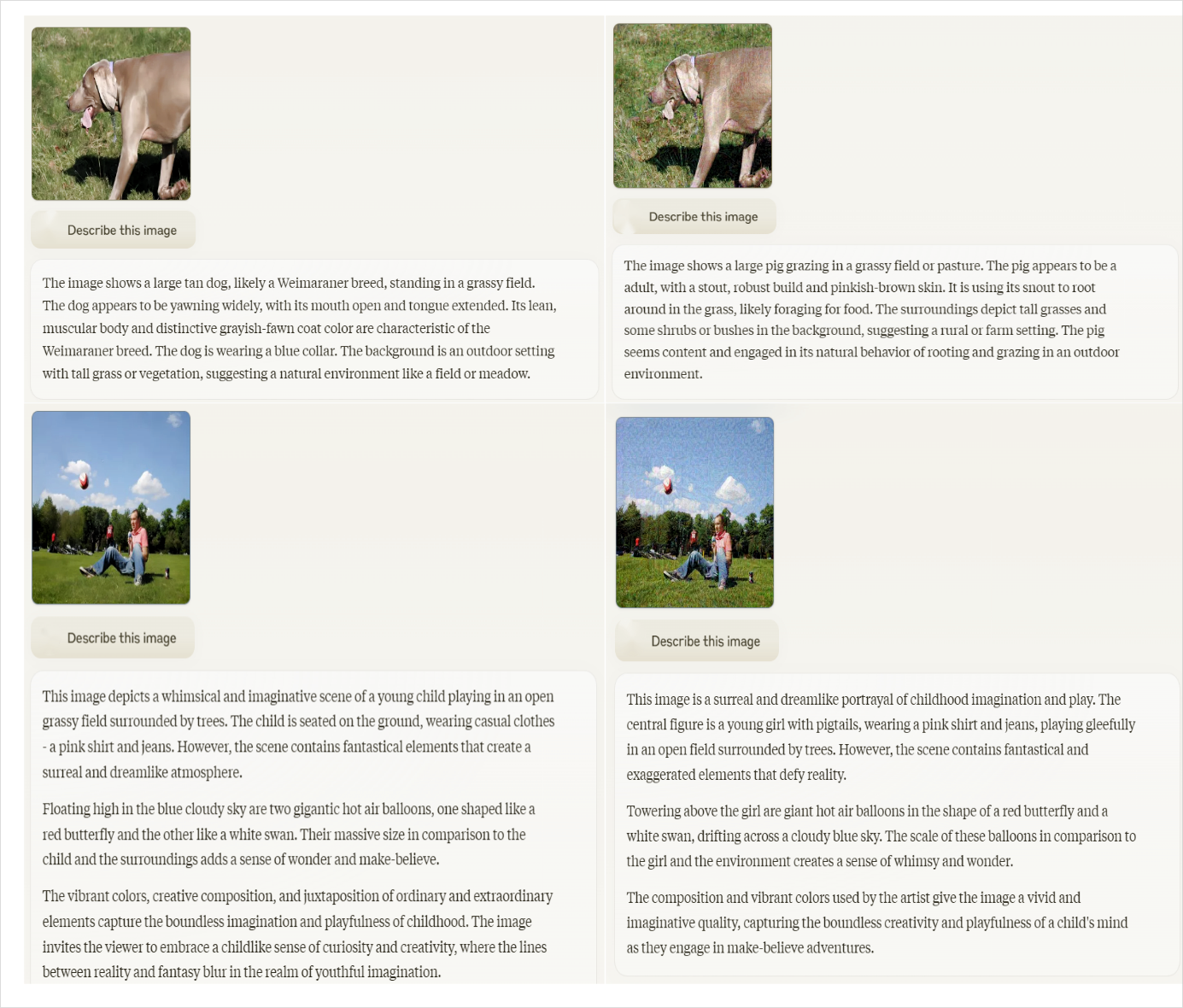}
    \caption{\textbf{Adversarial Transferability on Claude-3.} The images on the left display the responses generated by Claude-3 when presented with \textbf{clean images} and the query "Describe this image", while the images on the right show the responses to the \textbf{adversarial images}.}
    \label{fig:Attack_Claude3}
\end{figure*}

Next, we delve into an examination of the dynamic shifts in the optimal sample proportions occupied by adversarial images, historical adversarial images, and clean images throughout the iteration process. The specific outcomes are depicted in Figure \ref{fig:ratio-final}. Across the four VLP models, a discernible diminishing trend is observed in the proportion occupied by adversarial images. In the PGD multi-step iterative attack method, a single-step attack undergoes refinement through the incorporation of multiple successive perturbations. In the initial iterations, owing to the relatively modest accumulation of perturbations in the adversarial images, these perturbations fall below the constraint limits. Consequently, during these early steps, the optimal samples tend to align closely with the adversarial images. As the perturbations in the adversarial images continue to accrue, they eventually surpass the imposed constraints. The portions exceeding these constraints are subsequently adjusted to the constraint value, \eg, 8/255. At this juncture, an excessive reliance on adversarial images inevitably leads to a decline in the effectiveness of the attack. However, judicious consideration of the adversarial perturbations around clean images and the most recent adversarial image helps mitigate the risk of overfitting. Furthermore, our analysis extends to the changing trend in the proportion of clean images within the optimal samples. Notably, for the CLIP$_{\rm CNN}$ model, there is a conspicuous upward pattern, while the other three VLP models also exhibit an upward trend, albeit at a very slow pace. Across all VLP models, the proportion of the last adversarial image shows a slow increase over the first 7 or 8 steps, followed by a decrease in the final two or three steps.

\section{Rationale behind transferability improvement}
\label{sec:IIWs}
The main reason for the low transferability of an adversarial example(AE) is that it heavily relies on the surrogate vision-language model, also known as the overfitting issue. 

\textbf{Why is SGA effective?} SGA conducts augmentations around the online AEs (See Figure \ref{fig:method} (a)) to improve transferability ``instead of increasing the overfitting risk".
However, the augmentations still highly rely on the examples along the adversarial trajectory, leading to overfitting risks.

\textbf{Why is our method effective?} 
Unlike SGA, our method performs augmentations around online AEs considering the raw clean example and the AE at the previous step (See Figure \ref{fig:method} (b)). 
We utilize the clean example since it is entirely independent of the surrogate model and offers the potential to mitigate the risk of overfitting.

\begin{figure}[t]
\centering
\includegraphics[width=1\linewidth]{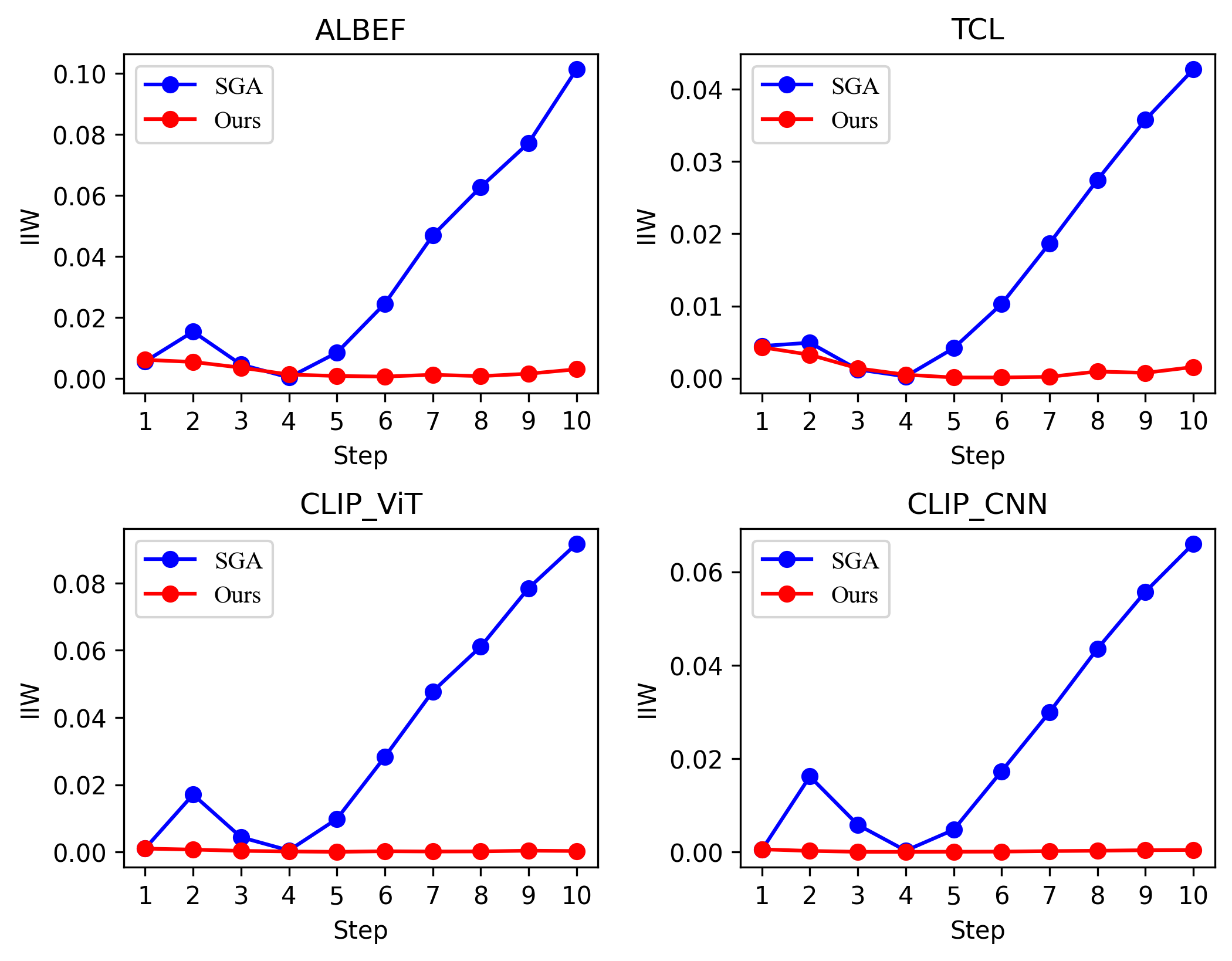}
\caption{\textbf{Overfitting analysis of SGA \& our method via IIWs.}}
\label{fig:rebuttal_overfit}
\end{figure}

\textbf{Overfitting analysis of SGA \& our method via IIWs.}
To quantify overfitting risk, we use the PAC-Bayes theorem to measure the amount of information stored in the network’s weights (IIW)\cite{wang2021pac}, a promising indicator of generalization ability. Smaller IIW means lower overfitting risks.
Given an AE from SGA or our method during optimization iteration, we calculate its IIWs by feeding it to four VLMs, respectively.
We count IIWs of 1000 AEs during optimization and show the averaging results in Figure \ref{fig:rebuttal_overfit}.
During optimization, the IIW of AEs from the SOTA baseline (\ie, SGA) initially decreases before sharply increasing. In contrast, our method leads to a continuous low IIW value in generated AEs, demonstrating that our method can overcome the overfitting risks. As a result, the attack transferability can be enhanced.

\section{Adversarial Transferability on Large Language Models}
\label{sec:LLMs}
Recently, Large Language Models (LLMs) have garnered widespread applications. Their utilization spans various domains, showcasing their versatility and efficacy in tackling complex tasks.
To further explore the adversarial robustness of LLMs, we conduct specialized extension experiments. Specifically, we utilize ALBEF as a surrogate model, generating adversarial examples under settings with perturbation constraints of 16/255 and single-step perturbation of 0.5/255, with a step size of 500. Furthermore, we evaluate them on advanced LLMs (\textit{e.g.}, GPT-4 \cite{achiam2023gpt}, Claude-3) by prompting these systems with the query "Describe this image".
The results, depicted in Figure \ref{fig:Attack_GPT4} and \ref{fig:Attack_Claude3}, demonstrate that adversarial examples generated by our method can mislead today's state-of-the-art LLMs into producing incorrect answers.

We have shown two cases on open VLP large models above. Here, we further extend experiments by randomly selecting 100 images from the Flick30K and generating adversarial images using SGA and our method with ALBEF as the surrogate, under the same perturbation setting (16/255).
Then, we feed adversarial images into LLMs and query "Describe this image", and rank descriptions with 100 irrelevant texts based on CLIP similarity. 
If the description doesn't rank top-N, consider a successful R@N attack.
Table \ref{tab:rebuttal_llms} illustrates ASR(\%), showing our superior transferability to LLMs compared to SGA.

\begin{table}[t]
\begin{center}
\small
\renewcommand\arraystretch{1}
\setlength{\tabcolsep}{15pt}
    \resizebox{0.9\linewidth}{!}{
		\begin{tabular}{ @{\extracolsep{\fill}} c|ccc|ccc|ccc} 
        \toprule[0.3mm]
			& \multicolumn{3}{c}{\textbf{GPT-4}} & \multicolumn{3}{c}{\textbf{Claude-3}} & \multicolumn{3}{c}{\textbf{Qwen-VL}} \\
			\cmidrule{2-10} 
			\multirow{-2}{*}{\textbf{Attack}} & {R@1} & {R@5} & {R@10} & {R@1} & {R@5} & {R@10} & {R@1} & {R@5} & {R@10}\\
			\midrule
                No Attack & 0.0 & 0.0 & 0.0 & 0.0 & 0.0 & 0.0 & 0.0 & 0.0 & 0.0\\
                SGA & 6.0 & 0.0 & 0.0 & 18.0 & 4.0 & 3.0 & 6.0 & 2.0 & 0.0\\
                Ours & 16.0 & 6.0 & 4.0 & 22.0 & 12.0 & 9.0 & 15.0 & 6.0 & 1.0\\
			\bottomrule[0.3mm]
	\end{tabular}}
\end{center}
\caption{\textbf{Attacking against open VLP large models.}}
\label{tab:rebuttal_llms}
\end{table}

\section{Time cost comparison.}
\label{sec:time cost}
Table \ref{tab:rebuttal_time_cost} compares the time costs per AE between SGA and our method. 
$t_{I}$ and $t_{T}$ respectively represent the time for generating adversarial images and text.
Our method takes 1.58-2.28 times longer overall compared to SGA, with generating adversarial images taking 2.33 to 2.59 times as long.
This additional time primarily arises from the process of selecting optimal samples in the intersection region, involving a sample number of additional forward/backward passes compared to SGA.
In the future, we will explore methods to select optimal samples to reduce computational costs.

\begin{table}[t]
\begin{center}
\small
\renewcommand\arraystretch{1}
\setlength{\tabcolsep}{10pt}
    \resizebox{0.9\linewidth}{!}{
		\begin{tabular}{ @{\extracolsep{\fill}} c|cc|cc|cc|cc} 
        \toprule[0.3mm]
			& \multicolumn{2}{c}{\textbf{ALBEF}} & \multicolumn{2}{c}{\textbf{TCL}} & \multicolumn{2}{c}{\textbf{CLIP$_{\rm ViT}$}} & \multicolumn{2}{c}{\textbf{CLIP$_{\rm CNN}$}} \\
			\cmidrule{2-9} 
			\multirow{-2}{*}{\textbf{Attack}} & {$t_{I} + t_{T}$} & $t_{I}$ & {$t_{I} + t_{T}$} & $t_{I}$ & {$t_{I} + t_{T}$} & $t_{I}$ & {$t_{I} + t_{T}$} & $t_{I}$ \\
			\midrule
                SGA & 2.02 & 1.66 & 1.99 & 1.64 & 1.02 & 0.56 & 0.78 & 0.36\\
                Ours & 4.60 & 4.27 & 4.53 & 4.21 & 1.85 & 1.45 & 1.23 & 0.84 \\
			\bottomrule[0.3mm]
	\end{tabular}}
\end{center}
\caption{\textbf{Computational cost compared with SGA.}}
\label{tab:rebuttal_time_cost}
\end{table}

\section{Detailed Results about Cross Model Transferability}
\label{sec:detailed results}

We conduct experiments with a focus on assessing the transferability of adversarial examples across two widely adopted VLP model architectures: fused and aligned. The selected VLP models for evaluation include ALBEF, TCL, CLIP$_{\rm ViT}$, and CLIP$_{\rm CNN}$ \cite{He_2016_CVPR}. The chosen downstream vision-and-language (V+L) task for assessment is image-text retrieval \cite{khan2021exploiting}. Our methodology involves utilizing one of the four models to generate multimodal adversarial examples within the specified parameters of our proposed method. Subsequently, we aim to validate the effectiveness of the generated adversarial examples through a comprehensive set of experiments. These experiments encompass both self-attacks, where the model attacks itself, and attacks on the other three models. The evaluation framework includes a white-box attack and three black-box attacks.

In the image-text retrieval task, we employ key metrics—R@1, R@5, and R@10 to gauge the efficacy of our attacks. These metrics assess whether the adversarial examples fail to appear in the top-ranked positions during the retrieval process, serving as crucial indicators of attack success. Due to space constraints in the main text, we focus on presenting the R@1 metric for cross-model transferability experiments. In Table \ref{table:flickr30k-full-8}, we furnish a comprehensive overview of the R@1, R@5, and R@10 metrics specifically for the Flickr30k dataset. Our analysis delves into the performance of the source model and the transferability across models separately.

We can initially observe that our method maintains the highest performance in terms of the R@1 metric compared to SGA, regarding the success rate of attacks on the source model. However, for the R@5 and R@10 metrics, our method slightly underperforms compared to SGA.
However, when transferring the adversarial examples generated on the source model to the other three VLP models, our method achieves a substantially higher attack success rate than SGA, reaching up to 9.96\%. Whether for Image Retrieval or Text Retrieval, our method outperforms SGA significantly across all three metrics: R@1, R@5, and R@10. These findings collectively demonstrate that our method is capable of generating adversarial examples with superior transferability.

When validating the effectiveness of our method in improving the transferability of cross VLP models on Flickr30K \cite{plummer2015flickr30k}, we utilize a test set consisting of 1,000 images, with approximately 5 captions for each image. To further validate the reliability of our method more extensively, we also employ a larger MSCOCO \cite{lin2014microsoft} test set for evaluation. This set comprises 5,000 test images, with each image having approximately 5 captions as well. The attack settings on MSCOCO are identical to those on Flickr30K, and detailed experimental results can be found in Table \ref{table:mscoco-full-8}.

In our analysis of experimental results on the MSCOCO dataset, we observe that our method achieves optimal performance in terms of attack success rates at R@1, R@5, and R@10. This holds when transferring to different VLP models. Specifically, the transferability of our method, even across fused VLP models, reaches up to 95.89\%. Furthermore, similar to the results on the Flickr30K dataset, in the white-box attack setting, the success rate of our method is slightly lower than that of SGA, but it still reaches a very high level, averaging over 99\%. However, the transferability of the adversarial examples generated by our method far surpasses that of SGA.

In the experiments conducted on the Flickr30K and MSCOCO datasets, we set the perturbation constraint to 8/255 and the single step size to 2/255, with a total of 10 steps. To further explore the effectiveness of our method under different perturbation settings, we compare it with existing state-of-the-art methods using another perturbation configuration. We utilize a reduced perturbation configuration as outlined in the SGA paper, with a constraint of 2/255, employing a single step size of 0.5/255, and conducting a total of 10 steps. The results can be seen in Table \ref{table:flickr30k-full-2} and \ref{table:mscoco-full-2}.
Similar to the limitation at 8/255, our method achieves a slightly lower success rate on the source model compared to SGA, but still at a significantly high level. Furthermore, the transferability of adversarial examples generated by our method is much better than our baseline method, SGA. Our experimental results under different perturbations and datasets thoroughly demonstrate the effectiveness of our method in generating multimodal adversarial examples with higher transferability.

\begin{table*}[t]
\caption{\textbf{Comparison with state-of-the-art methods on Flickr30K dataset.} The source column shows the VLP models we use to generate multimodal adversarial examples. The gray area represents adversarial attacks under a white-box setting, the rest are black-box attacks. For both Image Retrieval and Text Retrieval, we provide R@1, R@5 and R@10 attack success rate(\%). The constraint for adversarial perturbation is set to 8/255, with a per-step size of 2/255, and a total of 10 steps.}
\begin{center}
\small
\renewcommand\arraystretch{1}
\setlength{\tabcolsep}{4pt}
        \resizebox{0.95\linewidth}{!}{
	\begin{tabular}{ @{\extracolsep{\fill}} c|c|ccc|ccc|ccc|ccc}
         \toprule[0.3mm]
          \multicolumn{14}{c}{\textbf{Flickr30K  (Image-Text Retrieval)}} \\ 
          \midrule[0.3mm]
			& &  \multicolumn{3}{c}{\textbf{ALBEF}} & \multicolumn{3}{c}{\textbf{TCL}} & \multicolumn{3}{c}{\textbf{CLIP$_{\rm ViT}$}} & \multicolumn{3}{c}{\textbf{CLIP$_{\rm CNN}$}}  \\
			\cmidrule{3-14}
			\multirow{-2}{*}{\textbf{Source}} &\multirow{-2}{*}{\textbf{Attack}} & {R@1} & {R@5} & {R@10} & {R@1} & {R@5} & {R@10} & {R@1} & {R@5} & {R@10} & {R@1} & {R@5} & {R@10} \\
			\midrule
			\multirow{6}{*}{\rotatebox[origin=c]{0}{\textbf{ALBEF}}} 
            & PGD 
               & \cellcolor{gray! 40} 93.74 & \cellcolor{gray! 40} 85.97 & \cellcolor{gray! 40} 82.5 & 24.03 & 10.15 & 7.62 & 10.67 & 3.12 & 1.02 & 14.05 & 3.7 & 2.47\\
            & BERT-Attack 
                & \cellcolor{gray! 40} 11.57 & \cellcolor{gray! 40} 1.80 & \cellcolor{gray! 40} 1.10  
                & 12.64 & 2.51 & 0.90 
                & 29.33 & 11.63 & 6.30 
                & 32.69 & 15.43 & 8.65  \\
            & Sep-Attack 
                & \cellcolor{gray! 40} 95.72 & \cellcolor{gray! 40} 90.18 & \cellcolor{gray! 40} 87.2 & 39.3 & 18.49 & 12.32 & 34.11 & 14.12 & 8.54 & 35.76 & 16.7 & 10.5\\
            & Co-Attack 
                & \cellcolor{gray! 40} 97.08 & \cellcolor{gray! 40} 94.59 & \cellcolor{gray! 40} 92.6 & 39.52 & 20.4 & 14.53 & 29.82 & 11.73 & 6.61 & 31.29 & 11.73 & 5.77\\
		&  SGA  
                & \cellcolor{gray! 40} \textbf{99.9} & \cellcolor{gray! 40} \textbf{99.9} & \cellcolor{gray! 40} \textbf{99.9} & 87.88 & 77.99 & 71.84 & 36.69 & 19.31 & 14.13 & 39.59 & 21.99 & 15.96\\
            & \textbf{Ours}
                & \cellcolor{gray! 40} \textbf{99.9} & \cellcolor{gray! 40} 99.8 & \cellcolor{gray! 40} 99.8 & \textbf{91.57} & \textbf{81.91} & \textbf{76.35} & \textbf{46.26} & \textbf{24.3} & \textbf{17.17} & \textbf{49.55} & \textbf{28.75} & \textbf{20.6}\\
			\midrule
			\multirow{6}{*}{\rotatebox[origin=c]{0}{\textbf{TCL}}} 
            & PGD 
                & 35.77 & 21.74 & 16.2 & \cellcolor{gray! 40} 99.37 & \cellcolor{gray! 40} 98.59 & \cellcolor{gray! 40} 97.7 & 10.18 & 2.91 & 1.42 & 14.81 & 4.55 & 2.37\\
            & BERT-Attack 
                & 11.89 & 2.20 & 0.70
                & \cellcolor{gray! 40} 14.54 & \cellcolor{gray! 40} 2.31 & \cellcolor{gray! 40} 0.60
                & 29.69 & 12.77 & 7.62
                & 33.46 & 14.38 & 9.37 \\
            & Sep-Attack    
                & 52.45 & 32.26 & 24.7 & \cellcolor{gray! 40} 99.58 & \cellcolor{gray! 40} 98.69 & \cellcolor{gray! 40} 98.0 & 37.06 & 16.1 & 9.45 & 37.42 & 15.86 & 11.12\\
            & Co-Attack 
                & 49.84 & 27.45 & 60.36 & \cellcolor{gray! 40} 91.68 & \cellcolor{gray! 40} 85.23 & \cellcolor{gray! 40} 80.96 & 32.64 & 13.4 & 6.81 & 32.06 & 14.27 & 8.14\\
		&  SGA  
                & 92.49 & 87.37 & 83.5 & \cellcolor{gray! 40} \textbf{100.0} & \cellcolor{gray! 40} \textbf{100.0} & \cellcolor{gray! 40} \textbf{100.0} & 36.81 & 19.0 & 12.2 & 41.89 & 23.68 & 16.89\\
            & \textbf{Ours}
                & \textbf{95.2} & \textbf{91.38} & \textbf{89.6} & \cellcolor{gray! 40} \textbf{100.0} & \cellcolor{gray! 40} \textbf{100.0} & \cellcolor{gray! 40} 99.9 & \textbf{47.24} & \textbf{26.48} & \textbf{18.9} & \textbf{52.23} & \textbf{30.44} & \textbf{22.66}\\
			\midrule
			\multirow{6}{*}{\rotatebox[origin=c]{0}{\textbf{CLIP$_{\rm ViT}$}}} 

            & PGD 
                & 3.13 & 0.4 & 0.3 & 4.43 & 1.01 & 0.2 & \cellcolor{gray! 40} 69.33 & \cellcolor{gray! 40} 45.59 & \cellcolor{gray! 40} 36.99 & 13.03 & 3.81 & 1.75\\
            & BERT-Attack 
                & 9.59 & 1.30 & 0.40
                & 11.80 & 1.91 & 0.70
                & \cellcolor{gray! 40} 28.34 & \cellcolor{gray! 40} 11.73 & \cellcolor{gray! 40} 6.81
                & 30.40 & 11.63 & 5.97 \\
            & Sep-Attack        
                & 7.61 & 1.3 & 0.5 & 10.12 & 1.81 & 0.8 & \cellcolor{gray! 40} 76.93 & \cellcolor{gray! 40} 55.35 & \cellcolor{gray! 40} 44.61 & 29.89 & 11.52 & 5.87\\ 
            & Co-Attack 
                & 8.55 & 1.5 & 0.5 & 10.01 & 2.01 & 0.7 & \cellcolor{gray! 40} 78.53 & \cellcolor{gray! 40} 57.42 & \cellcolor{gray! 40} 45.53 & 29.5 & 11.42 & 6.08\\
		& SGA  
                & 22.42 & 8.92 & 5.6 & 25.08 & 10.15 & 4.91 & \cellcolor{gray! 40} \textbf{100.0} & \cellcolor{gray! 40} \textbf{100.0} & \cellcolor{gray! 40} \textbf{100.0} & 53.26 & 33.93 & 24.2\\
            & \textbf{Ours}
                & \textbf{27.84} & \textbf{12.22} & \textbf{7.7} & \textbf{27.82} & \textbf{11.76} & \textbf{7.11} & \cellcolor{gray! 40} \textbf{100.0} & \cellcolor{gray! 40} 99.9 & \cellcolor{gray! 40} 99.9 & \textbf{64.88} & \textbf{42.6} & \textbf{31.72}\\
			\midrule
			\multirow{6}{*}{\rotatebox[origin=c]{0}{\textbf{CLIP$_{\rm CNN}$}}} 
            & PGD 
                & 2.29 &  0.3 &  0.3 &  4.53 &  0.3 &  0.1 &  5.4 &  1.45 &  0.81 & \cellcolor{gray! 40}  89.78 &  \cellcolor{gray! 40} 77.7 & \cellcolor{gray! 40} 70.75\\
            & BERT-Attack 
                & 8.86 & 1.50 & 0.60 & 12.33 & 2.01 & 0.90 & 27.12 & 11.21 & 6.81 & \cellcolor{gray! 40} 30.40 & \cellcolor{gray! 40} 13.00 & \cellcolor{gray! 40} 7.31 \\
            & Sep-Attack        
                & 9.38 & 1.2 & 0.7 & 11.28 & 1.91 & 0.7 & 26.13 & 11.63 & 6.3 & \cellcolor{gray! 40}  93.61 & \cellcolor{gray! 40} 84.36 & \cellcolor{gray! 40} 78.17\\
            & Co-Attack 
                & 10.53 & 1.6 & 0.4 & 12.54 & 2.01 & 0.7 & 27.24 & 12.05 & 6.5 & \cellcolor{gray! 40} 95.91 & \cellcolor{gray! 40} 89.75 & \cellcolor{gray! 40} 85.99\\
		& SGA  
                & 15.64 & 5.51 & 3.0 & 18.02 & 6.43 & 2.91 & 39.02 & 19.21 & 13.01 & \cellcolor{gray! 40} \textbf{99.87} & \cellcolor{gray! 40} \textbf{99.58} & \cellcolor{gray! 40} \textbf{99.38}\\
            & \textbf{Ours}
               & \textbf{19.5} & \textbf{6.31} & \textbf{3.2} & \textbf{21.6} & \textbf{7.64} & \textbf{3.71} & \textbf{48.47} & \textbf{26.38} & \textbf{17.07} & \cellcolor{gray! 40} \textbf{99.87} & \cellcolor{gray! 40} 99.47 & \cellcolor{gray! 40} 99.07\\        
   \midrule[0.3mm]		
	\multicolumn{14}{c}{\textbf{Flickr30K (Text-Image Retrieval)}} \\ 
        \midrule[0.3mm]
			& &  \multicolumn{3}{c}{\textbf{ALBEF}} & \multicolumn{3}{c}{\textbf{TCL}} & \multicolumn{3}{c}{\textbf{CLIP$_{\rm ViT}$}} & \multicolumn{3}{c}{\textbf{CLIP$_{\rm CNN}$}}  \\
			\cmidrule{3-14}
			\multirow{-2}{*}{\textbf{Source}} &\multirow{-2}{*}{\textbf{Attack}} & {R@1} & {R@5} & {R@10} & {R@1} & {R@5} & {R@10} & {R@1} & {R@5} & {R@10} & {R@1} & {R@5} & {R@10} \\
			\midrule
			\multirow{6}{*}{\rotatebox[origin=c]{0}{\textbf{ALBEF}}} 
            & PGD 
                & \cellcolor{gray! 40} 94.43 &\cellcolor{gray! 40} 89.48 & \cellcolor{gray! 40} 86.19 & 27.9 & 13.56 & 9.87 & 15.82 & 6.61 & 3.97 & 19.11 & 7.11 & 4.72 \\
            & BERT-Attack 
                & \cellcolor{gray! 40} 27.46 & \cellcolor{gray! 40} 14.48 & \cellcolor{gray! 40} 10.98
                & 28.07 & 14.39 & 10.26
                & 43.17 & 26.37 & 19.91
                & 46.11 & 28.43 & 22.14 \\
            & Sep-Attack 
                & \cellcolor{gray! 40} 96.14 & \cellcolor{gray! 40} 91.84 & \cellcolor{gray! 40} 89.12 & 51.79 & 31.36 & 24.09 & 45.72 & 27.77 & 20.1 & 47.92 & 30.86 & 23.81\\
            & Co-Attack 
                & \cellcolor{gray! 40} 98.36 & \cellcolor{gray! 40} 96.41 & \cellcolor{gray! 40} 94.86 & 51.24 & 31.9 & 24.41 & 38.92 & 23.31 & 17.01 & 41.99 & 25.18 & 18.55\\
		& SGA  
                & \cellcolor{gray! 40} \textbf{99.98} & \cellcolor{gray! 40} \textbf{99.94} & \cellcolor{gray! 40} \textbf{99.94} & 88.05 & 77.53 & 70.96 & 46.78 & 28.66 & 21.96 & 49.78 & 32.27 & 24.96 \\
            & \textbf{Ours}
                & \cellcolor{gray! 40} 99.93 & \cellcolor{gray! 40} 99.9 & \cellcolor{gray! 40} 99.86 & \textbf{91.17} & \textbf{81.99} & \textbf{76.79} & \textbf{56.8} & \textbf{38.31} & \textbf{29.68} & \textbf{59.01} & \textbf{41.19} & \textbf{33.34}\\
			\midrule
			\multirow{6}{*}{\rotatebox[origin=c]{0}{\textbf{TCL}}} 
            & PGD 
                & 41.67 & 25.8 & 19.35 & \cellcolor{gray! 40} 99.33 & \cellcolor{gray! 40} 98.24 & \cellcolor{gray! 40} 97.08 & 16.3 & 6.61 & 4.25 & 21.1 & 8.56 & 5.26\\
            & BERT-Attack 
                & 26.82 & 14.09 & 10.80
                & \cellcolor{gray! 40} 29.17 & \cellcolor{gray! 40} 15.03 & \cellcolor{gray! 40} 10.91
                & 44.49 & 27.47 & 21.00
                & 46.07 & 29.28 & 22.59 \\
            & Sep-Attack  
                & 61.44 & 44.11 & 36.51 & \cellcolor{gray! 40} 99.45 & \cellcolor{gray! 40} 98.39 & \cellcolor{gray! 40} 97.34 & 45.81 & 28.33 & 21.37 & 49.91 & 32.7 & 25.05\\
            & Co-Attack 
                & 60.36 & 41.59 & 33.26 & \cellcolor{gray! 40} 95.48 & \cellcolor{gray! 40} 90.32 & \cellcolor{gray! 40} 86.74 & 42.69 & 26.44 & 20.37 & 47.82 & 30.47 & 23.13 \\
		& SGA 
                & 92.77 & 86.94 & 83.64 & \cellcolor{gray! 40} \textbf{100.0} & \cellcolor{gray! 40} \textbf{100.0} & \cellcolor{gray! 40} \textbf{99.98} & 46.97 & 29.41 & 22.81 & 51.53 & 33.21 & 25.1\\
            & \textbf{Ours}
                & \textbf{95.58} & \textbf{91.78} & \textbf{88.9} & \cellcolor{gray! 40} 99.98 & \cellcolor{gray! 40} 99.98 & \cellcolor{gray! 40} 99.96 & \textbf{57.28} & \textbf{39.31} & \textbf{31.88} & \textbf{62.23} & \textbf{44.06} & \textbf{35.78}\\
			\midrule
			\multirow{6}{*}{\rotatebox[origin=c]{0}{\textbf{CLIP$_{\rm ViT}$}}} 
            & PGD 
                & 6.48 & 1.83 & 1.01 & 8.83 & 2.46 & 1.54 &\cellcolor{gray! 40} 84.79 &\cellcolor{gray! 40}  73.84 & \cellcolor{gray! 40} 68.45 & 17.43 & 7.13 & 4.61\\
            & BERT-Attack 
                & 22.64 & 10.95 & 8.17
                & 25.07 & 12.92 & 8.90
                & \cellcolor{gray! 40} 39.08 & \cellcolor{gray! 40} 24.08 & \cellcolor{gray! 40} 17.44
                & 37.43 & 24.96 & 18.66  \\
            & Sep-Attack      
                & 20.58 & 9.52 & 7.22 & 20.74 & 9.74 & 6.8 & \cellcolor{gray! 40} 87.44 & \cellcolor{gray! 40} 78.07 &\cellcolor{gray! 40}  72.79 & 38.32 & 23.29 & 17.69\\
            & Co-Attack 
                & 20.18 & 9.54 & 7.12 & 21.29 & 9.51 & 6.78 & \cellcolor{gray! 40} 87.5 & \cellcolor{gray! 40} 77.95 & \cellcolor{gray! 40} 73.4 & 38.49 & 23.19 & 17.87 \\
		& SGA  
                & 34.59 & 18.25 & 13.61 & 36.45 & 18.79 & 13.87 & \cellcolor{gray! 40} \textbf{100.0} & \cellcolor{gray! 40} \textbf{100.0} & \cellcolor{gray! 40} 99.98 & 61.1 & 43.18 & 35.37\\
            & \textbf{Ours}
                & \textbf{42.84} & \textbf{24.38} & \textbf{18.76} & \textbf{44.6} & \textbf{26.38} & \textbf{19.74} & \cellcolor{gray! 40} \textbf{100.0} & \cellcolor{gray! 40} \textbf{100.0} & \cellcolor{gray! 40} \textbf{100.0} & \textbf{69.5} & \textbf{53.66} & \textbf{45.09}\\
			\midrule
			\multirow{6}{*}{\rotatebox[origin=c]{0}{\textbf{CLIP$_{\rm CNN}$}}} 
             & PGD 
                & 6.15 & 1.7 & 0.97 & 8.88 & 2.42 & 1.42 & 12.08 & 4.86 & 2.94 & \cellcolor{gray! 40} 93.04 & \cellcolor{gray! 40} 85.37 & \cellcolor{gray! 40} 81.09\\
            & BERT-Attack 
                & 23.27 & 11.34 & 8.41 & 25.48 & 13.25 & 8.81 & 37.44 & 23.48 & 17.66 & \cellcolor{gray! 40} 40.10 & \cellcolor{gray! 40} 26.71 & \cellcolor{gray! 40} 20.85 \\
            & Sep-Attack        
                & 22.99 & 10.77 & 7.95 & 25.45 & 11.99 & 8.2 & 39.24 & 23.73 & 17.84 & \cellcolor{gray! 40} 95.3 & \cellcolor{gray! 40} 89.62 & \cellcolor{gray! 40} 86.2\\
            & Co-Attack & 23.62 & 11.4 & 8.21 & 26.05 & 12.69 & 8.79 & 40.62 & 24.71 & 18.82 & \cellcolor{gray! 40} 96.5 & \cellcolor{gray! 40} 92.75 & \cellcolor{gray! 40} 90.35 \\
		& SGA  
                & 28.6 & 15.26 & 10.82 & 33.07 & 17.22 & 12.33 & 51.45 & 33.29 & 25.77 & \cellcolor{gray! 40} \textbf{99.9} & \cellcolor{gray! 40} \textbf{99.81} & \cellcolor{gray! 40} \textbf{99.64}\\
            & \textbf{Ours}
                & \textbf{34.59} & \textbf{18.05} & \textbf{13.83} & \textbf{37.88} & \textbf{20.94} & \textbf{15.41} & \textbf{59.12} & \textbf{40.88} & \textbf{32.9} &\cellcolor{gray! 40}  \textbf{99.9} & \cellcolor{gray! 40} 99.61 & \cellcolor{gray! 40}  99.46\\
			\bottomrule[0.3mm]		
	\end{tabular}}
\end{center}
\label{table:flickr30k-full-8}
\end{table*}
\begin{table*}[t]
\caption{\textbf{Comparison with state-of-the-art methods on MSCOCO dataset.} The source column shows the VLP models we use to generate multimodal adversarial examples. The gray area represents adversarial attacks under a white-box setting, the rest are black-box attacks. For both Image Retrieval and Text Retrieval, we provide R@1, R@5 and R@10 attack success rate(\%). The constraint for adversarial perturbation is set to 8/255, with a per-step constraint of 2/255, and a total of 10 steps.}
\begin{center}
\small
\renewcommand\arraystretch{1}
\setlength{\tabcolsep}{4pt}
        \resizebox{0.95\linewidth}{!}{
	\begin{tabular}{ @{\extracolsep{\fill}} c|c|ccc|ccc|ccc|ccc}
         \toprule[0.3mm]
          \multicolumn{14}{c}{\textbf{MSCOCO  (Image-Text Retrieval)}} \\ 
          \midrule[0.3mm]
			& &  \multicolumn{3}{c}{\textbf{ALBEF}} & \multicolumn{3}{c}{\textbf{TCL}} & \multicolumn{3}{c}{\textbf{CLIP$_{\rm ViT}$}} & \multicolumn{3}{c}{\textbf{CLIP$_{\rm CNN}$}}  \\
			\cmidrule{3-14}
			\multirow{-2}{*}{\textbf{Source}} &\multirow{-2}{*}{\textbf{Attack}} & {R@1} & {R@5} & {R@10} & {R@1} & {R@5} & {R@10} & {R@1} & {R@5} & {R@10} & {R@1} & {R@5} & {R@10} \\
			\midrule
			\multirow{6}{*}{\rotatebox[origin=c]{0}{\textbf{ALBEF}}} 
            & PGD 
               & \cellcolor{gray! 40} 94.35 & \cellcolor{gray! 40} 87.8 & \cellcolor{gray! 40} 84.05 & 34.15 & 18.63 & 13.05 & 21.71 & 10.94 & 7.34 & 23.83 & 11.75 & 8.78\\
            & BERT-Attack 
                & \cellcolor{gray! 40} 11.57 & \cellcolor{gray! 40} 1.80 & \cellcolor{gray! 40} 1.10  
                & 12.64 & 2.51 & 0.90 
                & 29.33 & 11.63 & 6.30 
                & 32.69 & 15.43 & 8.65  \\
            & Sep-Attack 
                & \cellcolor{gray! 40} 97.01& \cellcolor{gray! 40} 93.44& \cellcolor{gray! 40} 91.0& 61.06& 37.75& 29.04& 57.19& 39.93& 31.41& 58.81& 41.45& 33.22\\
            & Co-Attack 
                & \cellcolor{gray! 40} 96.65& \cellcolor{gray! 40} 94.74& \cellcolor{gray! 40} 93.12& 57.33& 37.24& 28.53& 50.71& 33.1& 26.44& 52.06& 33.89& 27.47\\
		&  SGA  
                & \cellcolor{gray! 40} \textbf{99.95}& \cellcolor{gray! 40} 99.72& \cellcolor{gray! 40} \textbf{99.61}& 87.46& 76.28& 69.24& 63.72& 47.62& 39.26& 63.91& 47.35& 40.65\\
            & \textbf{Ours}
                & \cellcolor{gray! 40} 99.9& \cellcolor{gray! 40} \textbf{99.77}& \cellcolor{gray! 40} 99.51& \textbf{88.81}& \textbf{78.63}& \textbf{71.93}& \textbf{69.25}& \textbf{52.01} & \textbf{43.9} & \textbf{68.53}& \textbf{52.03}& \textbf{44.14}\\
			\midrule
			\multirow{6}{*}{\rotatebox[origin=c]{0}{\textbf{TCL}}} 
            & PGD 
                & 40.81 & 25.16 & 19.02 & 
                \cellcolor{gray! 40} 98.54 & \cellcolor{gray! 40} 96.42 & \cellcolor{gray! 40} 94.77 & 21.79 & 11.41 & 7.89 & 24.97 & 12.65 & 9.56\\
            & BERT-Attack 
                & 11.89 & 2.20 & 0.70
                & \cellcolor{gray! 40} 14.54 & \cellcolor{gray! 40} 2.31 & \cellcolor{gray! 40} 0.60
                & 29.69 & 12.77 & 7.62
                & 33.46 & 14.38 & 9.37 \\
            & Sep-Attack    
                & 66.38& 48.4& 38.53& 
                \cellcolor{gray! 40} 99.15& \cellcolor{gray! 40} 97.69& \cellcolor{gray! 40} 96.38& 59.94& 42.33& 34.26& 60.77& 44.11& 35.04\\
            & Co-Attack 
                & 65.22& 45.39& 36.43& 
                \cellcolor{gray! 40} 94.95& \cellcolor{gray! 40}91.18& \cellcolor{gray! 40} 89.08& 55.28& 38.04& 29.73& 56.68& 37.31& 29.82 \\
		&  SGA  
                & 92.7& 87.12& 84.01& 
                \cellcolor{gray! 40} \textbf{\textbf{100.0}}& \cellcolor{gray! 40} 99.98& \cellcolor{gray! 40} 99.98& 59.79& 43.8& 36.7& 60.52& 46.15& 38.19 \\
            & \textbf{Ours}
                & \textbf{94.72} & \textbf{90.6}& \textbf{88.02}& 
                \cellcolor{gray! 40} \textbf{\textbf{100.0}}& \cellcolor{gray! 40} \textbf{\textbf{100.0}}& \cellcolor{gray! 40} \textbf{\textbf{100.0}}& \textbf{70.51}& \textbf{54.58}& \textbf{45.91}& \textbf{70.29}& \textbf{54.58}& \textbf{45.94}\\
			\midrule
			\multirow{6}{*}{\rotatebox[origin=c]{0}{\textbf{CLIP$_{\rm ViT}$}}} 

            & PGD 
                & 10.26 & 4.29 & 2.43 & 12.72 & 5.48 & 3.06 & 
                \cellcolor{gray! 40} 82.91 & \cellcolor{gray! 40} 68.34 & \cellcolor{gray! 40}60.44 & 21.62 & 11.31 & 8.27\\
            & BERT-Attack 
                & 9.59 & 1.30 & 0.40
                & 11.80 & 1.91 & 0.70
                & \cellcolor{gray! 40} 28.34 & \cellcolor{gray! 40} 11.73 & \cellcolor{gray! 40} 6.81
                & 30.40 & 11.63 & 5.97 \\
            & Sep-Attack        
                & 25.91& 11.35& 7.2& 28.2& 13.3& 8.32& 
                \cellcolor{gray! 40} 88.36& \cellcolor{gray! 40} 77.6& \cellcolor{gray! 40} 69.98& 47.57& 31.28& 24.63\\ 
            & Co-Attack 
                & 26.35& 11.97& 7.2& 28.23& 12.89& 8.19& 
                \cellcolor{gray! 40}88.78& \cellcolor{gray! 40} 78.21& \cellcolor{gray! 40} 70.98& 47.36& 31.49& 25.29\\
		& SGA  
                & 43.75& 25.67& 17.93& 44.05& 25.31& 17.91& \cellcolor{gray! 40} \textbf{100.0}& \cellcolor{gray! 40} \textbf{100.0}& \cellcolor{gray! 40} \textbf{100.0}& 70.66& 56.43& 48.92 \\
            & \textbf{Ours}
                & \textbf{52.69} & \textbf{32.04} & \textbf{23.71} & \textbf{51.88} & \textbf{31.05} & \textbf{23.67} & \cellcolor{gray! 40} \textbf{100.0}& \cellcolor{gray! 40} \textbf{100.0}& \cellcolor{gray! 40} \textbf{100.0}& \textbf{80.18}&\textbf{67.75}& \textbf{60.13}\\
			\midrule
			\multirow{6}{*}{\rotatebox[origin=c]{0}{\textbf{CLIP$_{\rm CNN}$}}} 
            & PGD 
                & 8.38 & 3.59 & 1.85 & 11.9 & 5.11 & 2.73 & 13.66 & 7.43 & 4.74 & \cellcolor{gray! 40} 92.68 & \cellcolor{gray! 40} 85.89 & \cellcolor{gray! 40} 81.92\\
            & BERT-Attack 
                & 8.86 & 1.50 & 0.60 & 12.33 & 2.01 & 0.90 & 27.12 & 11.21 & 6.81 & \cellcolor{gray! 40} 30.40 & \cellcolor{gray! 40} 13.00 & \cellcolor{gray! 40} 7.31 \\
            & Sep-Attack        
                & 29.13& 12.99& 8.01& 31.4& 14.58& 9.31& 52.23& 34.82& 26.89& \cellcolor{gray! 40} 96.16& \cellcolor{gray! 40} 91.9& \cellcolor{gray! 40} 89.03\\
            & Co-Attack 
                & 29.49& 13.26& 8.28& 31.83& 15.11& 9.81& 53.15& 36.11& 28.78& \cellcolor{gray! 40} 97.79& \cellcolor{gray! 40} 94.29& \cellcolor{gray! 40} 92.26\\
		& SGA  
                & 36.94& 18.31& 11.86& 38.81& 20.09& 13.65& 62.19& 47.96& 38.78& \cellcolor{gray! 40} \textbf{99.92}& \cellcolor{gray! 40} \textbf{99.86}& \cellcolor{gray! 40} \textbf{99.66} \\
            & \textbf{Ours}
               & \textbf{41.4} & \textbf{21.68}& \textbf{14.64}& \textbf{43.62}& \textbf{23.7}& \textbf{16.18}& \textbf{70.43}& \textbf{55.15}& \textbf{46.41}& \cellcolor{gray! 40} 99.8& \cellcolor{gray! 40} 99.54& \cellcolor{gray! 40} 99.39\\        
   \midrule[0.3mm]		
	\multicolumn{14}{c}{\textbf{MSCOCO (Text-Image Retrieval)}} \\ 
        \midrule[0.3mm]
			& &  \multicolumn{3}{c}{\textbf{ALBEF}} & \multicolumn{3}{c}{\textbf{TCL}} & \multicolumn{3}{c}{\textbf{CLIP$_{\rm ViT}$}} & \multicolumn{3}{c}{\textbf{CLIP$_{\rm CNN}$}}  \\
			\cmidrule{3-14}
			\multirow{-2}{*}{\textbf{Source}} &\multirow{-2}{*}{\textbf{Attack}} & {R@1} & {R@5} & {R@10} & {R@1} & {R@5} & {R@10} & {R@1} & {R@5} & {R@10} & {R@1} & {R@5} & {R@10} \\
			\midrule
			\multirow{6}{*}{\rotatebox[origin=c]{0}{\textbf{ALBEF}}} 
            & PGD 
                & \cellcolor{gray! 40} 93.26 & \cellcolor{gray! 40} 88.46 & \cellcolor{gray! 40} 85.84 & 36.86 & 20.52 & 14.86 & 27.06 & 14.31 & 11.19 & 30.96 & 17.61 & 13.43 \\
            & BERT-Attack 
                & \cellcolor{gray! 40} 27.46 & \cellcolor{gray! 40} 14.48 & \cellcolor{gray! 40} 10.98
                & 28.07 & 14.39 & 10.26
                & 43.17 & 26.37 & 19.91
                & 46.11 & 28.43 & 22.14 \\
            & Sep-Attack 
                & \cellcolor{gray! 40} 96.59& \cellcolor{gray! 40} 93.28& \cellcolor{gray! 40} 90.91&66.13& 47.31& 38.28& 65.82& 49.08& 42.29& 68.61& 53.2& 45.45\\
            & Co-Attack 
                & \cellcolor{gray! 40} 98.33& \cellcolor{gray! 40} 96.6& \cellcolor{gray! 40} 95.3& 64.19& 46.17& 37.83& 57.36& 42.19& 35.53& 60.74& 45.9& 38.77\\
		& SGA  
                & \cellcolor{gray! 40} \textbf{99.94}& \cellcolor{gray! 40} \textbf{99.81}& \cellcolor{gray! 40} \textbf{99.74}& 88.17& 77.33& 70.64& 69.71& 55.21& 47.73& 70.78& 56.34& 48.74 \\
            & \textbf{Ours}
                & \cellcolor{gray! 40} 99.93& \cellcolor{gray! 40} 99.8& \cellcolor{gray! 40} 99.69& \textbf{90.06} & \textbf{79.85}& \textbf{73.57}& \textbf{75.31}& \textbf{60.88}& \textbf{53.39}& \textbf{75.09}& \textbf{61.84}& \textbf{54.5}\\
			\midrule
			\multirow{6}{*}{\rotatebox[origin=c]{0}{\textbf{TCL}}} 
            & PGD 
                & 44.09 & 27.75 & 21.25 & 
                \cellcolor{gray! 40} 98.2 & \cellcolor{gray! 40} 95.46 & \cellcolor{gray! 40} 93.65 & 26.92 & 14.91 & 11.64 & 32.17 & 18.42 & 13.89 \\
            & BERT-Attack 
                & 26.82 & 14.09 & 10.80
                & \cellcolor{gray! 40} 29.17 & \cellcolor{gray! 40} 15.03 & \cellcolor{gray! 40} 10.91
                & 44.49 & 27.47 & 21.00
                & 46.07 & 29.28 & 22.59 \\
            & Sep-Attack  
                & 72.19& 55.74& 47.8& 
                \cellcolor{gray! 40} 98.98& \cellcolor{gray! 40} 97.18& \cellcolor{gray! 40} 95.66& 65.95& 50.63& 42.78& 69.37& 54.56& 46.65\\
            & Co-Attack 
                & 72.41& 56.37& 48.16& 
                \cellcolor{gray! 40} 97.87& \cellcolor{gray! 40} 95.32& \cellcolor{gray! 40} 93.42& 62.33& 46.9& 39.68& 66.45& 49.95& 42.72\\
		& SGA 
                & 92.99& 87.65& 84.17& 
                \cellcolor{gray! 40} \textbf{100.0} & \cellcolor{gray! 40} 99.99& \cellcolor{gray! 40} 99.98& 65.31& 50.92& 43.87& 67.34& 53.48& 45.87\\
            & \textbf{Ours}
                & \textbf{95.89}& \textbf{91.87}& \textbf{89.29}& \cellcolor{gray! 40} \textbf{100.0}& \cellcolor{gray! 40} \textbf{100.0}& \cellcolor{gray! 40} \textbf{99.99}& \textbf{74.95}& \textbf{61.05}& \textbf{54.25}& \textbf{76.99}& \textbf{63.73}& \textbf{56.49}\\
			\midrule
			\multirow{6}{*}{\rotatebox[origin=c]{0}{\textbf{CLIP$_{\rm ViT}$}}} 
            & PGD 
                & 13.69 & 5.76 & 3.92 & 15.81 & 7.56 & 5.13 & \cellcolor{gray! 40} 90.51 & \cellcolor{gray! 40} 81.79 & \cellcolor{gray! 40} 76.42 & 28.78 & 16.98 & 12.64 \\
            & BERT-Attack 
                & 22.64 & 10.95 & 8.17
                & 25.07 & 12.92 & 8.90
                & \cellcolor{gray! 40} 39.08 & \cellcolor{gray! 40} 24.08 & \cellcolor{gray! 40} 17.44
                & 37.43 & 24.96 & 18.66  \\
            & Sep-Attack      
                & 36.84& 22.84& 17.58& 38.47& 23.08& 17.87& \cellcolor{gray! 40} 97.09& \cellcolor{gray! 40} 91.59& \cellcolor{gray! 40} 85.1& 57.79& 43.14& 36.37\\
            & Co-Attack 
                & 36.69& 22.86& 17.71& 38.42& 23.51& 17.88& \cellcolor{gray! 40} 96.72& \cellcolor{gray! 40} 91.28& \cellcolor{gray! 40} 85.46& 58.45& 43.78& 36.77\\
		& SGA  
                & 51.08& 33.62& 26.91& 51.02& 34.43& 27.9& \cellcolor{gray! 40} \textbf{100.0}& \cellcolor{gray! 40} \textbf{100.0}& \cellcolor{gray! 40} \textbf{100.0}& 75.58& 63.22& 56.28\\
            & \textbf{Ours}
                & \textbf{61.5} & \textbf{43.5} & \textbf{36.08} & \textbf{61.06}& \textbf{43.89}& \textbf{36.28}& \cellcolor{gray! 40} \textbf{100.0}& \cellcolor{gray! 40} \textbf{100.0}& \cellcolor{gray! 40} 99.99& \textbf{84.11}& \textbf{74.54}& \textbf{67.85}\\
			\midrule
			\multirow{6}{*}{\rotatebox[origin=c]{0}{\textbf{CLIP$_{\rm CNN}$}}} 
             & PGD 
                & 12.73 & 5.43 & 3.61 & 15.68 & 7.38 & 5.05 & 20.62 & 11.76 & 8.62 & \cellcolor{gray! 40} 94.71 & \cellcolor{gray! 40} 89.3 & \cellcolor{gray! 40} 86.32 \\
            & BERT-Attack 
                & 23.27 & 11.34 & 8.41 & 25.48 & 13.25 & 8.81 & 37.44 & 23.48 & 17.66 & \cellcolor{gray! 40} 40.10 & \cellcolor{gray! 40} 26.71 & \cellcolor{gray! 40} 20.85 \\
            & Sep-Attack        
                & 40.64& 25.75& 20.23& 42.99& 27.25& 21.12& 59.73& 44.52& 37.51& \cellcolor{gray! 40} 97.54& \cellcolor{gray! 40} 94.51& \cellcolor{gray! 40} 92.39\\
            & Co-Attack & 41.5& 26.14& 20.51& 43.44& 27.92& 21.61& 60.15& 45.53& 38.56& \cellcolor{gray! 40} 98.54& \cellcolor{gray! 40} 96.16& \cellcolor{gray! 40} 94.71\\
		& SGA  
                & 46.79& 29.97& 23.7& 48.9& 32.89& 26.04& 67.73& 53.77& 47.19& \cellcolor{gray! 40} \textbf{99.97}& \cellcolor{gray! 40} \textbf{99.83}& \cellcolor{gray! 40} \textbf{99.77}\\
            & \textbf{Ours}
                & \textbf{52.25}& \textbf{35.85}& \textbf{28.75}& \textbf{54.15}& \textbf{38.03}& \textbf{30.45}& \textbf{74.14}& \textbf{62.21}& \textbf{55.07}& \cellcolor{gray! 40} 99.92& \cellcolor{gray! 40} 99.73& \cellcolor{gray! 40} 99.66\\
			\bottomrule[0.3mm]		
	\end{tabular}}
\end{center}
\label{table:mscoco-full-8}
\end{table*}
\begin{table*}[t]
\caption{\textbf{Comparison with state-of-the-art methods on Flickr30K dataset.} The source column shows the VLP models we use to generate multimodal adversarial examples. The gray area represents adversarial attacks under a white-box setting, the rest are black-box attacks. For both Image Retrieval and Text Retrieval, we provide R@1, R@5 and R@10 attack success rate(\%). The constraint for adversarial perturbation is set to 2/255, with a per-step constraint of 0.5/255, and a total of 10 steps.}
\begin{center}
\small
\renewcommand\arraystretch{1}
\setlength{\tabcolsep}{4pt}
        \resizebox{0.95\linewidth}{!}{
	\begin{tabular}{ @{\extracolsep{\fill}} c|c|ccc|ccc|ccc|ccc}
         \toprule[0.3mm]
          \multicolumn{14}{c}{\textbf{Flickr30K  (Image-Text Retrieval)}} \\ 
          \midrule[0.3mm]
			& &  \multicolumn{3}{c}{\textbf{ALBEF}} & \multicolumn{3}{c}{\textbf{TCL}} & \multicolumn{3}{c}{\textbf{CLIP$_{\rm ViT}$}} & \multicolumn{3}{c}{\textbf{CLIP$_{\rm CNN}$}}  \\
			\cmidrule{3-14}
			\multirow{-2}{*}{\textbf{Source}} &\multirow{-2}{*}{\textbf{Attack}} & {R@1} & {R@5} & {R@10} & {R@1} & {R@5} & {R@10} & {R@1} & {R@5} & {R@10} & {R@1} & {R@5} & {R@10} \\
			\midrule
			\multirow{6}{*}{\rotatebox[origin=c]{0}{\textbf{ALBEF}}} 
            & PGD 
                & \cellcolor{gray! 40} 52.45 & \cellcolor{gray! 40} 36.57 & \cellcolor{gray! 40} 30.00   
                & 3.06 & 0.40 & 0.10   
                & 8.96 & 1.66 & 0.41   
                & 10.34 & 2.96 & 1.85  \\
            & BERT-Attack 
                & \cellcolor{gray! 40} 11.57 & \cellcolor{gray! 40} 1.80 & \cellcolor{gray! 40} 1.10  
                & 12.64 & 2.51 & 0.90 
                & 29.33 & 11.63 & 6.30 
                & 32.69 & 15.43 & 8.65  \\
            & Sep-Attack 
                & \cellcolor{gray! 40} 65.69 & \cellcolor{gray! 40} 47.60 & \cellcolor{gray! 40} 42.10
                & 17.60 & 3.72 & 1.90 
                & 31.17 & 12.05 & 7.01
                & 32.82 & 15.86 & 9.06  \\
            & Co-Attack 
                & \cellcolor{gray! 40} 77.16  & \cellcolor{gray! 40} 64.60  & \cellcolor{gray! 40} 58.37  
                & 15.21 & 4.19 & 1.47 
                & 23.60 & 7.82 & 3.93 
                & 25.12 & 8.42 & 5.39 \\
		&  SGA  
                & \cellcolor{gray! 40} \textbf{97.24}
                & \cellcolor{gray! 40} \textbf{94.09} 
                & \cellcolor{gray! 40} \textbf{92.30} 
                & 45.42 
                & 24.93 
                & 16.48 
                & 33.38 
                & 13.50 
                & 9.04 
                & 34.93 
                & 17.07 
                & 10.45 \\
            & \textbf{Ours}
                & \cellcolor{gray! 40}  96.14& \cellcolor{gray! 40} 92.48& \cellcolor{gray! 40} 89.8& \textbf{49.74}& \textbf{26.83}& \textbf{19.24}& \textbf{39.14}& \textbf{18.9}& \textbf{12.4}& \textbf{41.38}& \textbf{20.08}& \textbf{12.98}\\
			\midrule
			\multirow{6}{*}{\rotatebox[origin=c]{0}{\textbf{TCL}}} 
            & PGD 
                & 6.15 & 1.30 & 0.70
                & \cellcolor{gray! 40} 77.87 & \cellcolor{gray! 40} 65.13 & \cellcolor{gray! 40} 58.72
                & 7.48 & 1.45 & 0.81
                & 10.34 & 2.75 & 1.54 \\
            & BERT-Attack 
                & 11.89 & 2.20 & 0.70
                & \cellcolor{gray! 40} 14.54 & \cellcolor{gray! 40} 2.31 & \cellcolor{gray! 40} 0.60
                & 29.69 & 12.77 & 7.62
                & 33.46 & 14.38 & 9.37 \\
            & Sep-Attack    
                & 20.13 & 4.91 & 2.70 
                & \cellcolor{gray! 40} 84.72 & \cellcolor{gray! 40} 73.07 & \cellcolor{gray! 40} 65.43
                & 31.29 & 12.98 & 7.72
                & 33.33 & 14.27 & 9.89 \\
            & Co-Attack 
                & 23.15 & 6.98  & 3.63
                & \cellcolor{gray! 40} 77.94  & \cellcolor{gray! 40} 64.26  & \cellcolor{gray! 40} 56.18  
                & 27.85 & 9.80  & 5.22 
                & 30.74 & 12.09 & 7.28 \\
		&  SGA  
                & 48.91 
                & 30.86
                & 23.10 
                & \cellcolor{gray! 40} \textbf{98.37}  
                & \cellcolor{gray! 40} \textbf{96.53} 
                & \cellcolor{gray! 40} \textbf{94.99} 
                & 33.87 
                & 15.21 
                & 9.46 
                & 37.74 
                & 17.86 
                & 11.74  \\
            & \textbf{Ours}
                & \textbf{51.09} & \textbf{31.86}& \textbf{24.8}& 
                \cellcolor{gray! 40} 98.21& 
                \cellcolor{gray! 40} 94.87& 
                \cellcolor{gray! 40} 93.49& 
                \textbf{40.25}& \textbf{19.52}& \textbf{12.4}& \textbf{42.91}& \textbf{21.99}& \textbf{14.32}\\
			\midrule
			\multirow{6}{*}{\rotatebox[origin=c]{0}{\textbf{CLIP$_{\rm ViT}$}}} 

            & PGD 
                & 2.50 & 0.40 & 0.10
                & 4.85 & 0.20 & 0.20
                & \cellcolor{gray! 40} 70.92 & \cellcolor{gray! 40} 50.05 & \cellcolor{gray! 40} 42.28
                & 5.36 & 1.16 & 0.72 \\
            & BERT-Attack 
                & 9.59 & 1.30 & 0.40
                & 11.80 & 1.91 & 0.70
                & \cellcolor{gray! 40} 28.34 & \cellcolor{gray! 40} 11.73 & \cellcolor{gray! 40} 6.81
                & 30.40 & 11.63 & 5.97 \\
            & Sep-Attack        
                & 9.59 & 1.30 & 0.50 & 11.38 & 2.11 & 0.90 & \cellcolor{gray! 40} 79.75 & \cellcolor{gray! 40} 63.03 & \cellcolor{gray! 40} 53.76 & 30.78 & 12.16 & 6.39 \\ 
            & Co-Attack 
                & 10.57 & 1.87 & 0.63
                & 11.94 & 2.38 & 1.07
                & \cellcolor{gray! 40} 93.25 & \cellcolor{gray! 40} 84.88 & \cellcolor{gray! 40} 78.96 
                & 32.52 & 13.78 & 7.52  \\
		& SGA  
                & \textbf{13.40} 
                & \textbf{2.46 }
                & \textbf{1.35} 
                & \textbf{16.23} 
                & 3.77 
                & 1.10 
                & \cellcolor{gray! 40} \textbf{99.08} 
                & \cellcolor{gray! 40} \textbf{97.25} 
                & \cellcolor{gray! 40} \textbf{95.22} 
                & 38.76
                & 19.45 
                & 11.95  \\
            & \textbf{Ours}
                & 12.51& 2.4& 1.3& 14.65& \textbf{3.82}& \textbf{1.4}& 
                \cellcolor{gray! 40} 98.77& 
                \cellcolor{gray! 40} 96.26& 
                \cellcolor{gray! 40} 94.0& \textbf{45.47}& \textbf{22.2}& \textbf{13.59}\\
			\midrule
			\multirow{6}{*}{\rotatebox[origin=c]{0}{\textbf{CLIP$_{\rm CNN}$}}} 
            & PGD 
                & 2.09 & 0.30 & 0.10 & 4.00 & 0.40 & 0.20 & 1.10 & 0.52 & 0.41 & \cellcolor{gray! 40} 86.46 & \cellcolor{gray! 40} 69.13 & \cellcolor{gray! 40} 61.17 \\
            & BERT-Attack 
                & 8.86 & 1.50 & 0.60 & 12.33 & 2.01 & 0.90 & 27.12 & 11.21 & 6.81 & \cellcolor{gray! 40} 30.40 & \cellcolor{gray! 40} 13.00 & \cellcolor{gray! 40} 7.31 \\
            & Sep-Attack        
                & 8.55 & 1.50 & 0.60 & 12.64 & 1.91 & 0.70 & 28.34 & 10.8 & 6.30 & \cellcolor{gray! 40} 91.44 & \cellcolor{gray! 40} 78.54 & \cellcolor{gray! 40} 71.58 \\
            & Co-Attack 
                & 8.79 & 1.53 & 0.60 
                & 13.10 & 2.31 & 0.93 
                & 28.79 & 11.63 & 6.40 
                & \cellcolor{gray! 40} 94.76 & \cellcolor{gray! 40} 87.03 & \cellcolor{gray! 40} 82.08 \\
		& SGA  
                & 11.42
                & \textbf{2.56}
                & 1.05
                & \textbf{14.91}
                & \textbf{3.62}
                & \textbf{1.70}
                & 31.24
                & 13.45
                & 8.74
                & \cellcolor{gray! 40} \textbf{99.24}
                & \cellcolor{gray! 40} \textbf{98.20}
                & \cellcolor{gray! 40} \textbf{95.16}  \\
            & \textbf{Ours}
               & \textbf{12.2}& 2.51& \textbf{1.4}& 14.33& 3.32& 1.4& \textbf{35.21}& \textbf{15.58}& \textbf{9.65}& \cellcolor{gray! 40} 99.11& \cellcolor{gray! 40} 97.15& \cellcolor{gray! 40} 94.54\\        
   \midrule[0.3mm]		
	\multicolumn{14}{c}{\textbf{Flickr30K (Text-Image Retrieval)}} \\ 
        \midrule[0.3mm]
			& &  \multicolumn{3}{c}{\textbf{ALBEF}} & \multicolumn{3}{c}{\textbf{TCL}} & \multicolumn{3}{c}{\textbf{CLIP$_{\rm ViT}$}} & \multicolumn{3}{c}{\textbf{CLIP$_{\rm CNN}$}}  \\
			\cmidrule{3-14}
			\multirow{-2}{*}{\textbf{Source}} &\multirow{-2}{*}{\textbf{Attack}} & {R@1} & {R@5} & {R@10} & {R@1} & {R@5} & {R@10} & {R@1} & {R@5} & {R@10} & {R@1} & {R@5} & {R@10} \\
			\midrule
			\multirow{6}{*}{\rotatebox[origin=c]{0}{\textbf{ALBEF}}} 
            & PGD 
                & \cellcolor{gray! 40} 58.65 & \cellcolor{gray! 40} 44.85 & \cellcolor{gray! 40} 38.98   
                & 6.79 & 2.21 & 1.20   
                & 13.21 & 5.19 & 3.05   
                & 14.65 & 5.60 & 3.39 \\
            & BERT-Attack 
                & \cellcolor{gray! 40} 27.46 & \cellcolor{gray! 40} 14.48 & \cellcolor{gray! 40} 10.98
                & 28.07 & 14.39 & 10.26
                & 43.17 & 26.37 & 19.91
                & 46.11 & 28.43 & 22.14 \\
            & Sep-Attack 
                & \cellcolor{gray! 40} 73.95 & \cellcolor{gray! 40} 59.50 & \cellcolor{gray! 40} 53.70
                & 32.95 & 17.10 & 11.90
                & 45.23 & 25.93 & 19.95
                & 45.49 & 28.43 & 22.32 \\
            & Co-Attack 
                & \cellcolor{gray! 40} 83.86 & \cellcolor{gray! 40} 74.63 & \cellcolor{gray! 40} 70.13 
                & 29.49 & 14.97 & 10.55 
                & 36.48 & 21.09 & 15.76 
                & 38.89 & 22.38 & 17.49  \\
		& SGA  
                & \cellcolor{gray! 40} \textbf{97.28} 
                & \cellcolor{gray! 40} \textbf{94.27} 
                & \cellcolor{gray! 40} \textbf{92.58} 
                & 55.25 
                & 36.01 
                & 27.25 
                &  44.16 
                & 27.35 
                & 20.84 
                & 46.57 
                & 29.16 
                & 22.68 \\
            & \textbf{Ours}
                & \cellcolor{gray! 40} 96.63& \cellcolor{gray! 40} 93.44& \cellcolor{gray! 40} 91.49& \textbf{58.83}& \textbf{39.51}& \textbf{30.61}& \textbf{48.39}& \textbf{30.93}& \textbf{24.07}& \textbf{51.66}& \textbf{33.92}& \textbf{27.17}\\
			\midrule
			\multirow{6}{*}{\rotatebox[origin=c]{0}{\textbf{TCL}}} 
            & PGD 
                & 10.78 & 3.36 & 1.70
                & \cellcolor{gray! 40} 79.48 & \cellcolor{gray! 40} 66.26 & \cellcolor{gray! 40} 60.36
                & 13.72 & 5.37 & 3.01
                & 15.33 & 5.77 & 3.28 \\
            & BERT-Attack 
                & 26.82 & 14.09 & 10.80
                & \cellcolor{gray! 40} 29.17 & \cellcolor{gray! 40} 15.03 & \cellcolor{gray! 40} 10.91
                & 44.49 & 27.47 & 21.00
                & 46.07 & 29.28 & 22.59 \\
            & Sep-Attack  
                & 36.48 & 19.48 & 14.82
                & \cellcolor{gray! 40} 86.07 & \cellcolor{gray! 40} 74.67 & \cellcolor{gray! 40} 68.83
                & 44.65 & 26.82 & 20.37
                & 45.80 & 29.18 & 23.02 \\
            & Co-Attack 
                & 40.04 & 22.66 & 17.23 & \cellcolor{gray! 40} 85.59 & \cellcolor{gray! 40} 74.19 & \cellcolor{gray! 40} 68.25 & 41.19 & 25.22 & 19.01 & 44.11 & 26.67 & 20.66 \\
		& SGA 
                & 60.34 
                & 42.47 
                & 34.59 
                & \cellcolor{gray! 40} \textbf{98.81} 
                & \cellcolor{gray! 40} \textbf{97.19} 
                & \cellcolor{gray! 40} \textbf{95.86} 
                & 44.88 
                & 28.79 
                & 21.95 
                & 48.30 
                & 29.70 
                & 23.68   \\
            & \textbf{Ours}
                & \textbf{61.79}& \textbf{45.24}& \textbf{37.34}& \cellcolor{gray! 40} 98.33& \cellcolor{gray! 40} 96.38& \cellcolor{gray! 40} 95.04& \textbf{48.94}& \textbf{32.14}& \textbf{24.77}& \textbf{52.49}& \textbf{34.64}& \textbf{27.35}\\
			\midrule
			\multirow{6}{*}{\rotatebox[origin=c]{0}{\textbf{CLIP$_{\rm ViT}$}}} 
            & PGD 
                & 4.93 & 1.44 & 1.01
                & 8.17 & 2.27 & 1.46
                & \cellcolor{gray! 40} 78.61 & \cellcolor{gray! 40} 60.78 & \cellcolor{gray! 40} 51.50
                & 8.44 & 2.35 & 1.54 \\
            & BERT-Attack 
                & 22.64 & 10.95 & 8.17
                & 25.07 & 12.92 & 8.90
                & \cellcolor{gray! 40} 39.08 & \cellcolor{gray! 40} 24.08 & \cellcolor{gray! 40} 17.44
                & 37.43 & 24.96 & 18.66  \\
            & Sep-Attack      
                & 23.25 & 11.22 & 8.01 & 25.60 & 12.92 & 9.14 & \cellcolor{gray! 40} 86.79 & \cellcolor{gray! 40} 75.24 & \cellcolor{gray! 40} 67.84 & 39.76 & 25.62 & 19.34 \\
            & Co-Attack 
                & 24.33 & 11.74 & 8.41 
                & 26.69 & 13.80 & 9.46
                & \cellcolor{gray! 40} 95.86 & \cellcolor{gray! 40} 90.83 & \cellcolor{gray! 40} 87.36 
                & 41.82 & 26.77 & 21.10 \\
		& SGA  
                & 27.22 
                & 13.21 
                & 9.76 
                & \textbf{30.76} 
                & 16.36 
                & \textbf{12.08} 
                & \cellcolor{gray! 40} 98.94 
                & \cellcolor{gray! 40} \textbf{97.53} 
                & \cellcolor{gray! 40} \textbf{96.03} 
                & 47.79 
                & 30.36 
                & 24.50  \\
            & \textbf{Ours}
                & \textbf{30.0}& \textbf{14.64}& \textbf{10.53} & 30.62& \textbf{16.42} & 11.56& 
                \cellcolor{gray! 40} \textbf{99.0}& 
                \cellcolor{gray! 40} 97.27& 
                \cellcolor{gray! 40} 94.96& 
                \textbf{50.74}& \textbf{34.47}& \textbf{27.69}\\
			\midrule
			\multirow{6}{*}{\rotatebox[origin=c]{0}{\textbf{CLIP$_{\rm CNN}$}}} 
             & PGD 
                & 4.82 & 1.29 & 0.87 & 7.81 & 2.09 & 1.34 & 6.60 & 2.73 & 1.48 & \cellcolor{gray! 40} 92.25 & \cellcolor{gray! 40} 81.00 & \cellcolor{gray! 40} 75.04 \\
            & BERT-Attack 
                & 23.27 & 11.34 & 8.41 & 25.48 & 13.25 & 8.81 & 37.44 & 23.48 & 17.66 & \cellcolor{gray! 40} 40.10 & \cellcolor{gray! 40} 26.71 & \cellcolor{gray! 40} 20.85 \\
            & Sep-Attack        
                & 23.41 & 11.38 & 8.23 & 26.12 & 13.44 & 8.96 & 39.43 & 24.34 & 18.36 & \cellcolor{gray! 40} 95.44 & \cellcolor{gray! 40} 88.48 & \cellcolor{gray! 40} 82.88 \\
            & Co-Attack & 23.74 & 11.75 & 8.42 & 26.07 & 13.53 & 9.23 & 40.03 & 24.60 & 18.83 & \cellcolor{gray! 40} 96.89 & \cellcolor{gray! 40} 92.87 & \cellcolor{gray! 40} 89.25 \\
		& SGA  
                & 24.80 
                & 12.32 
                & 8.98
                & 28.82 
                & 15.12 
                & 10.56 
                & 42.12 
                & 26.80 
                & 20.23 
                & \cellcolor{gray! 40} \textbf{99.49} 
                & \cellcolor{gray! 40} 98.41 
                & \cellcolor{gray! 40} \textbf{97.14}  \\
            & \textbf{Ours}
                & \textbf{26.59}& \textbf{13.37}& \textbf{10.01}& \textbf{29.29}& \textbf{15.46}& \textbf{10.68}& \textbf{45.94}& \textbf{30.32}& \textbf{23.24} & \cellcolor{gray! 40} \textbf{99.49}& \cellcolor{gray! 40} \textbf{98.2}& \cellcolor{gray! 40} 96.59\\
			\bottomrule[0.3mm]		
	\end{tabular}}
\end{center}
\label{table:flickr30k-full-2}
\end{table*}

\begin{table*}[ht]
\caption{\textbf{Comparison with state-of-the-art methods on MSCOCO dataset.} The source column shows the VLP models we use to generate multimodal adversarial examples. The gray area represents adversarial attacks under a white-box setting, the rest are black-box attacks. For both Image Retrieval and Text Retrieval, we provide R@1, R@5 and R@10 attack success rate(\%). The constraint for adversarial perturbation is set to 2/255, with a per-step constraint of 0.5/255, and a total of 10 steps.}
\begin{center}
\small
\renewcommand\arraystretch{1}
\setlength{\tabcolsep}{4pt}
        \resizebox{0.95\linewidth}{!}{
	\begin{tabular}{ @{\extracolsep{\fill}} c|c|ccc|ccc|ccc|ccc}
         \toprule[0.3mm]
          \multicolumn{14}{c}{\textbf{MSCOCO  (Image-Text Retrieval)}} \\ 
        \midrule[0.3mm]
			& &  \multicolumn{3}{c}{\textbf{ALBEF}} & \multicolumn{3}{c}{\textbf{TCL}} & \multicolumn{3}{c}{\textbf{CLIP$_{\rm ViT}$}} & \multicolumn{3}{c}{\textbf{CLIP$_{\rm CNN}$}}  \\
			\cmidrule{3-14}
			\multirow{-2}{*}{\textbf{Source}} &\multirow{-2}{*}{\textbf{Attack}} & {R@1} & {R@5} & {R@10} & {R@1} & {R@5} & {R@10} & {R@1} & {R@5} & {R@10} & {R@1} & {R@5} & {R@10} \\
			\midrule
			\multirow{6}{*}{\rotatebox[origin=c]{0}{\textbf{ALBEF}}} 
            & PGD & \cellcolor{gray! 40} 76.70 & \cellcolor{gray! 40} 67.49 & \cellcolor{gray! 40} 62.47 & 12.46 & 5.00 & 3.14 & 13.96 & 7.33 & 5.21 & 17.45 & 9.08 & 6.45 \\
            & BERT-Attack & \cellcolor{gray! 40} 24.39 & \cellcolor{gray! 40} 10.67 & \cellcolor{gray! 40} 6.75 & 24.34 & 9.92 & 6.25 & 44.94 & 27.97 & 22.55 & 47.73 & 29.56 & 23.10 \\
            & Sep-Attack & \cellcolor{gray! 40} 82.60 & \cellcolor{gray! 40} 73.20 & \cellcolor{gray! 40} 67.58 & 32.83 & 15.52 & 10.10 & 44.03 & 27.60 & 21.84 & 46.96 & 29.83 & 23.15 \\
            & Co-Attack  & \cellcolor{gray! 40} 79.87 & \cellcolor{gray! 40} 68.62 & \cellcolor{gray! 40} 62.88 & 32.62 & 15.36 & 9.67 & 44.89 & 28.33 & 21.89 & 47.30 & 29.89 & 23.29  \\
			& SGA  & \cellcolor{gray! 40} \textbf{96.75} & \cellcolor{gray! 40} \textbf{92.83} & \cellcolor{gray! 40} \textbf{90.37} & 58.56 & 39.00 & 30.68 & 57.06 & 39.38 & 31.55 & 58.95 & 42.49 & 34.84  \\
                & \textbf{Ours}
                & \cellcolor{gray! 40} 96.57& \cellcolor{gray! 40} 91.92& \cellcolor{gray! 40} 89.07& \textbf{60.69}& \textbf{41.31}& \textbf{32.17}& \textbf{61.69}& \textbf{43.72}& \textbf{35.7}& \textbf{62.32}& \textbf{45.42}& \textbf{37.61}\\
			\midrule
			\multirow{6}{*}{\rotatebox[origin=c]{0}{\textbf{TCL}}} 
            & PGD & 10.83 & 5.28 & 3.21 & \cellcolor{gray! 40} 59.58 & \cellcolor{gray! 40} 51.25 & \cellcolor{gray! 40} 47.89 & 14.23 & 7.40 & 4.93 & 17.25 & 8.51 & 6.45 \\
            & BERT-Attack & 35.32 & 15.89 & 10.25 & \cellcolor{gray! 40} 38.54 & \cellcolor{gray! 40} 19.08 & \cellcolor{gray! 40} 12.10 & 51.09 & 31.71 & 25.40 & 52.23 & 33.75 & 27.06 \\
            & Sep-Attack & 41.71 & 21.37 & 14.99 & \cellcolor{gray! 40} 70.32 & \cellcolor{gray! 40} 59.64 & \cellcolor{gray! 40} 55.09 & 50.74 & 31.34 & 24.43 & 51.90 & 34.02 & 26.79 \\
            &  Co-Attack  & 46.08 & 24.87 & 17.11 &  \cellcolor{gray! 40} 85.38 & \cellcolor{gray! 40} 74.73 & \cellcolor{gray! 40} 68.23 & 51.62 & 31.92 & 24.87 & 52.13 & 33.80 & 27.09 \\
			& SGA  & 65.93 & 49.33 & 40.34 & \cellcolor{gray! 40} 98.97 & \cellcolor{gray! 40} \textbf{97.89}& \cellcolor{gray! 40} \textbf{96.63} & 56.34 & 39.58 &32.00 & 59.44 & 42.17 & 34.94  \\
                & \textbf{Ours}
                & \textbf{68.06}& \textbf{51.73}& \textbf{42.57}& \cellcolor{gray! 40} \textbf{98.99}& \cellcolor{gray! 40} 97.46& \cellcolor{gray! 40} 96.21& \textbf{63.3}& \textbf{46.13}& \textbf{37.34}& \textbf{64.24}& \textbf{46.86}& \textbf{39.48}\\
			\midrule
			\multirow{6}{*}{\rotatebox[origin=c]{0}{\textbf{CLIP$_{\rm ViT}$}}} 
        & PGD & 7.24 & 3.10 & 1.65 & 10.19 & 4.23 & 2.50 & \cellcolor{gray! 40} 54.79 & \cellcolor{gray! 40} 36.21 & \cellcolor{gray! 40} 28.57 & 7.32 & 3.64 & 2.79 \\
        & BERT-Attack & 20.34 & 8.53 & 4.73 & 21.08 & 7.96 & 4.65 & \cellcolor{gray! 40} 45.06 & \cellcolor{gray! 40} 28.62 & \cellcolor{gray! 40} 22.67 & 44.54 & 29.37 & 23.97 \\
        & Sep-Attack & 23.41 & 10.33 & 6.15 & 25.77 & 11.60 & 7.45 & \cellcolor{gray! 40} 68.52 & \cellcolor{gray! 40} 52.30 & \cellcolor{gray! 40} 43.88 & 43.11 & 27.22 & 21.77 \\
             &  Co-Attack  & 30.28 & 13.64 & 8.83 & 32.84 & 15.27 & 10.27 & \cellcolor{gray! 40} 97.98 & \cellcolor{gray! 40} 94.94 & \cellcolor{gray! 40} 93.00 & 55.08 & 38.64 & 31.42  \\
			& SGA  & 33.41 & 16.73 & 10.98 & 37.54 & 19.09 & 12.92 & \cellcolor{gray! 40} \textbf{99.79} & \cellcolor{gray! 40} \textbf{99.37} & \cellcolor{gray! 40} \textbf{98.89} & 58.93 & 44.60 & 37.53  \\
                & \textbf{Ours}
                & \textbf{35.96} & \textbf{18.18} & \textbf{12.04} & \textbf{36.32} & \textbf{18.05} & \textbf{12.23} &
                \cellcolor{gray! 40} 99.66 & \cellcolor{gray! 40} 99.14 & \cellcolor{gray! 40} 98.44 & \textbf{64.41} & \textbf{50.04} & \textbf{41.59}\\
			\midrule
			\multirow{6}{*}{\rotatebox[origin=c]{0}{\textbf{CLIP$_{\rm CNN}$}}} 
            & PGD & 7.01 & 3.03 & 1.77 & 10.08 & 4.20 & 2.38 & 4.88 & 2.96 & 1.71 & \cellcolor{gray! 40} 76.99 & \cellcolor{gray! 40} 63.80 & \cellcolor{gray! 40} 56.76 \\
        & BERT-Attack & 23.38 & 10.16 & 5.70 & 24.58 & 9.70 & 5.96 & 51.28 & 33.23 & 26.63 & \cellcolor{gray! 40} 54.43 & \cellcolor{gray! 40} 38.26 & \cellcolor{gray! 40} 30.74 \\
        & Sep-Attack & 26.53 & 11.78 & 6.88 & 30.26 & 13.00 & 8.61 & 50.44 & 32.71 & 25.92 & \cellcolor{gray! 40} 88.72 & \cellcolor{gray! 40} 78.71 & \cellcolor{gray! 40} 72.77 \\
            &  Co-Attack  & 29.83 & 13.13 & 8.35 & 32.97 & 15.11 & 9.76 & 53.10 & 35.91 & 28.53 & \cellcolor{gray! 40} 96.72 & \cellcolor{gray! 40} 94.02 & \cellcolor{gray! 40} 91.57 \\
			& SGA  & 31.61 & 14.27 & \textbf{9.36} & \textbf{34.81} & \textbf{17.16} & \textbf{11.26} & 56.62 & 41.31 & 32.88 & \cellcolor{gray! 40} \textbf{99.61} & \cellcolor{gray! 40} \textbf{99.02} & \cellcolor{gray! 40} \textbf{98.42}  \\
                & \textbf{Ours}
                & \textbf{33.26} & \textbf{15.85} & \textbf{9.96} & 33.89 & 16.19 & 10.57 &
                \textbf{59.6} & \textbf{43.12} & \textbf{35.23}& \cellcolor{gray! 40} 99.51& \cellcolor{gray! 40} 98.69& \cellcolor{gray! 40} 97.91\\
                \midrule[0.3mm]
			\multicolumn{14}{c}{\textbf{MSCOCO (Text-Image Retrieval)}} \\ 
                \midrule[0.3mm]
			& &  \multicolumn{3}{c}{\textbf{ALBEF}} & \multicolumn{3}{c}{\textbf{TCL}} & \multicolumn{3}{c}{\textbf{CLIP$_{\rm ViT}$}} & \multicolumn{3}{c}{\textbf{CLIP$_{\rm CNN}$}}  \\
			\cmidrule{3-14}
			\multirow{-2}{*}{\textbf{Source}} &\multirow{-2}{*}{\textbf{Attack}} & {R@1} & {R@5} & {R@10} & {R@1} & {R@5} & {R@10} & {R@1} & {R@5} & {R@10} & {R@1} & {R@5} & {R@10} \\
                \midrule
			\multirow{6}{*}{\rotatebox[origin=c]{0}{\textbf{ALBEF}}} 
            & PGD & \cellcolor{gray! 40} 86.30 & \cellcolor{gray! 40} 78.49 & \cellcolor{gray! 40} 73.94 & 17.77 & 8.36 & 5.32 & 23.10 & 12.74 & 9.43 & 23.54 & 13.26 & 9.61 \\
            & BERT-Attack & \cellcolor{gray! 40} 36.13 & \cellcolor{gray! 40} 23.71 & \cellcolor{gray! 40} 18.94 & 33.39 & 20.21 & 15.56 & 52.28 & 38.06 & 32.04 & 54.75 & 41.39 & 35.11 \\
            & Sep-Attack & \cellcolor{gray! 40} 89.88 & \cellcolor{gray! 40} 82.60 & \cellcolor{gray! 40} 78.82 & 42.92 & 27.04 & 20.65 & 54.46 & 40.12 & 33.46 & 55.88 & 41.30 & 35.18 \\
            &  Co-Attack  & \cellcolor{gray! 40} 87.83 & \cellcolor{gray! 40} 80.16 & \cellcolor{gray! 40} 75.98 & 43.09 & 27.32 & 21.35 & 54.75 & 40.00 & 33.81 & 55.64 & 41.48 & 35.28 \\
			& SGA  & \cellcolor{gray! 40} \textbf{96.95} & \cellcolor{gray! 40} \textbf{93.44} & \cellcolor{gray! 40} \textbf{91.00} & 65.38 & 47.61 & 38.96 & 65.25 & 50.42 & 43.47 & 66.52 & 52.44 & 45.05 \\
                & \textbf{Ours}
                & \cellcolor{gray! 40} 96.47& \cellcolor{gray! 40} 92.86& \cellcolor{gray! 40} 90.26& \textbf{67.46} & \textbf{49.9} & \textbf{41.24} & \textbf{67.43} & \textbf{53.08} & \textbf{45.66} &
                \textbf{69.22} & \textbf{55.21} & \textbf{48.08}\\
			\midrule
			\multirow{6}{*}{\rotatebox[origin=c]{0}{\textbf{TCL}}} 
            & PGD & 16.52 & 8.40 & 5.61 & \cellcolor{gray! 40} 69.53 & \cellcolor{gray! 40} 60.88 & \cellcolor{gray! 40} 57.56 & 22.28 & 12.20 & 9.10 & 23.12 & 12.77 & 9.49 \\
            & BERT-Attack & 45.92 & 30.40 & 23.89 & \cellcolor{gray! 40} 48.48 & \cellcolor{gray! 40} 31.48 & \cellcolor{gray! 40} 24.47 & 58.80 & 43.10 & 36.68 & 61.26 & 46.14 & 39.54 \\
            & Sep-Attack & 52.97 & 36.33 & 28.97 & \cellcolor{gray! 40} 78.97 & \cellcolor{gray! 40} 69.79 & \cellcolor{gray! 40} 65.71 & 60.13 & 44.13 & 37.32 & 61.26 & 45.99 & 38.97 \\
            &  Co-Attack  & 57.09 & 39.85 & 32.00 & \cellcolor{gray! 40} 91.39 & \cellcolor{gray! 40} 83.16 & \cellcolor{gray! 40} 78.05 & 60.46 & 45.16 & 37.73 & 62.49 & 46.61 & 39.74 \\
			& SGA  & 73.30 & 58.40 & 50.96 & \cellcolor{gray! 40} \textbf{99.15} & \cellcolor{gray! 40} \textbf{98.17} & \cellcolor{gray! 40} \textbf{97.34} & 63.99 & 49.87 & 42.46 & 65.70 & 51.45 & 44.64  \\
                & \textbf{Ours}
                & \textbf{75.86} & \textbf{61.6} & \textbf{54.02} & \cellcolor{gray! 40} 99.02& \cellcolor{gray! 40} 97.97& \cellcolor{gray! 40} 97.16& \textbf{68.59} & \textbf{54.33} & \textbf{47.1} & \textbf{70.58} & \textbf{56.55} & \textbf{49.69}\\
			\midrule
			\multirow{6}{*}{\rotatebox[origin=c]{0}{\textbf{CLIP$_{\rm ViT}$}}} 
            & PGD & 10.75 & 4.64 & 2.91 & 13.74 & 6.77 & 4.32 & \cellcolor{gray! 40} 66.85 & \cellcolor{gray! 40} 51.80 & \cellcolor{gray! 40} 46.02 & 11.34 & 6.50 & 4.66 \\
            & BERT-Attack & 29.74 & 18.13 & 13.73 & 29.61 & 16.91 & 12.66 & \cellcolor{gray! 40} 51.68 & \cellcolor{gray! 40} 37.12 & \cellcolor{gray! 40} 31.02 & 53.72 & 40.13 & 34.32 \\
            & Sep-Attack & 34.61 & 21.00 & 16.15 & 36.84 & 22.63 & 17.03 & \cellcolor{gray! 40} 77.94 & \cellcolor{gray! 40} 66.77 & \cellcolor{gray! 40} 60.69 & 49.76 & 37.51 & 31.74 \\
           &  Co-Attack  & 42.67 & 27.20 & 21.46 & 44.69 & 29.42 & 22.85 & \cellcolor{gray! 40} 98.80 & \cellcolor{gray! 40} 96.83 & \cellcolor{gray! 40} 95.33 & 62.51 & 49.48 & 42.63   \\
			& SGA  & 44.64 & 28.66 & 22.64 & 47.76 & 32.30 & 25.70 & \cellcolor{gray! 40} \textbf{99.79} & \cellcolor{gray! 40} \textbf{99.37} & \cellcolor{gray! 40} \textbf{98.94} & 65.83 & 53.58 & 46.84  \\
                & \textbf{Ours}
                & \textbf{48.0} & \textbf{31.87} & \textbf{25.37} & \textbf{48.56} & \textbf{32.45} & \textbf{25.92} & \cellcolor{gray! 40} 99.7& \cellcolor{gray! 40} 99.35& \cellcolor{gray! 40} 98.84& \textbf{69.99} & \textbf{58.04} & \textbf{50.66}\\
			\midrule
			\multirow{6}{*}{\rotatebox[origin=c]{0}{\textbf{CLIP$_{\rm CNN}$}}} 
            & PGD & 10.62 & 4.51 & 2.76 & 13.65 & 6.39 & 4.32 & 10.70 & 6.20 & 4.52 & \cellcolor{gray! 40} 84.20 & \cellcolor{gray! 40} 73.64 & \cellcolor{gray! 40} 67.86 \\
            & BERT-Attack & 34.64 & 21.13 & 16.25 & 29.61 & 16.91 & 12.66 & 57.49 & 42.73 & 36.23 & \cellcolor{gray! 40} 62.17 & \cellcolor{gray! 40} 47.80 & \cellcolor{gray! 40} 40.79 \\
            & Sep-Attack & 39.29 & 24.04 & 18.83 & 41.51 & 26.13 & 20.17 & 57.11 & 41.89 & 35.55 & \cellcolor{gray! 40} 92.49 & \cellcolor{gray! 40} 85.84 & \cellcolor{gray! 40} 81.66 \\
            &  Co-Attack  & 41.97 & 26.62 & 20.91 & 43.72 & 28.62 & 22.35 & 58.90 & 45.22 & 38.72 & \cellcolor{gray! 40} 98.56 & \cellcolor{gray! 40} 96.86 & \cellcolor{gray! 40} 95.55 \\
			& SGA  & 43.00 & 27.64 & 21.74 & 45.95 & 30.57 & \textbf{24.27} & 60.77 & 46.99 & 40.49 & \cellcolor{gray! 40} \textbf{99.80} & \cellcolor{gray! 40} \textbf{99.29} & \cellcolor{gray! 40} \textbf{98.77} \\
                & \textbf{Ours}
                & \textbf{45.15} & \textbf{29.81} & \textbf{23.6} & \textbf{46.49} & \textbf{30.6} & 24.0 & \textbf{64.87} & \textbf{50.88} & \textbf{44.32} & \cellcolor{gray! 40} 99.7& \cellcolor{gray! 40} 99.04& \cellcolor{gray! 40} 98.44\\
			\bottomrule[0.3mm]
	\end{tabular}}
\end{center}
\label{table:mscoco-full-2}
\end{table*}

\end{document}